\documentclass{article}
\usepackage{iclr2027_conference,times}
\usepackage{amsmath,amssymb,bm}
\usepackage{graphicx,booktabs,array,longtable,float}
\usepackage{todonotes}
\usepackage{CJKutf8}
\usepackage{xcolor,microtype,url,hyperref,placeins,needspace}
\hypersetup{colorlinks=true,citecolor=blue,linkcolor=blue,urlcolor=blue}
\title{

Boosting Metric Depth Completion via Training-Free Adaptive Response Geometry

}
\author{Mia Zhang$^{1,2}$\\
$^{1}$dConstruct Robotics\\
$^{2}$University of Waterloo\\
\texttt{mia.zhang@dconstruct.it}
\And
Jizong Peng$^{1}$\\
$^{1}$dConstruct Robotics\\
\texttt{jizong.peng@dconstruct.ai}
}
\newcommand{\R}{\mathbb{R}}

\iclrfinalcopy
\begin{document}
\maketitle
\lhead{}
\raggedbottom
\setlength{\textfloatsep}{12pt plus 2pt minus 2pt}
\setlength{\floatsep}{10pt plus 2pt minus 2pt}
\setlength{\intextsep}{10pt plus 2pt minus 2pt}

\begin{abstract}
Depth completion aims to recover dense metric depth from sparse sensor measurements, increasingly leveraging visual foundation models as geometric priors. However, aligning these priors to true metric scale typically relies on rigid affine assumptions in predefined coordinate systems, leaving systematic calibration errors. Linearity in depth calibration depends on the response coordinate. We introduce adaptive response geometry, which makes the fixed choice of depth, log depth, or disparity an image-level unknown. A continuous response family unifies these coordinates and defines an explicit depth-dependent gain. We derive the response-gradient relation and estimate the response parameters in metric space. Hard-Dirichlet residual reconstruction completes the calibrated prior. Under deliberately incomplete metric observations, the training-free pipeline achieves macro AbsRel $0.0301$ and macro NMed $14.04^\circ$, improving both aggregate measures over PriorDA, LDCM, and Any2Full. Linearity diagnostics examine how the selected response changes the depth relation and its metric error.
\end{abstract}

\section{Introduction}
Monocular depth estimation provides a dense description of scene geometry from a single image.
Recent depth foundation models are trained on millions of images, including pseudo-labeled real images~\citep{yang2024depthv2,sekmen2026lumon}.
Their predictions provide geometric priors for reconstruction and robot perception, but need not represent distances in meters: monocular scale ambiguity, and potentially an unresolved shift, separate relative structure from metric geometry.

Monocular models represent and supervise relative depth in a chosen coordinate with affine or scale normalization.
MiDaS~\citep{ranftl2022towards} and Depth Anything~\citep{yang2024depth} use affine-invariant disparity space, where distant depths are compressed toward zero and errors on nearby surfaces receive greater emphasis.
ZoeDepth~\citep{bhat2023zoedepth} and Metric3D~\citep{yin2023metric3d} use scale-invariant supervision in log-depth, where equal depth ratios produce equal changes.
MoGe-3~\citep{kong2026moge3} aligns its point map by a scale and a shift along the optical axis, corresponding to an affine relation in metric depth.
These choices facilitate learning across heterogeneous datasets while prescribing the coordinate of linear alignment.

Depth completion uses sparse LiDAR or incomplete RGB-D observations to resolve depth scale ambiguities.
Each observed pixel pairs a prior value with a physical measurement, constraining a conversion that must also apply where measurements are absent.
\begin{figure}[!ht]
\centering
\includegraphics[width=\textwidth]{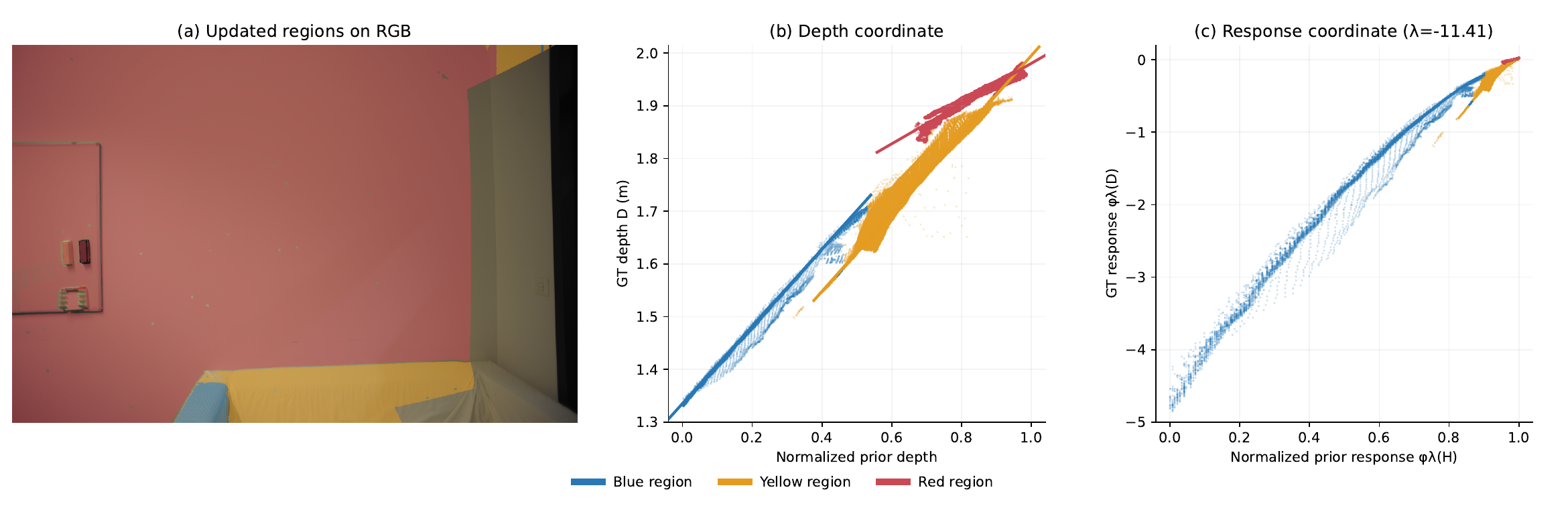}
\caption{Depth-dependent alignment differences can be alleviated by adapting a single global response coordinate, without assigning separate affine parameters to individual regions.
(a) Colors indicate diagnostic point groups initialized from depth-percentile seeds and refined along the point bands, not semantic segments.
(b) RANSAC reference lines show their different local slopes and intercepts.
(c) The same points in the response coordinate with $\lambda=-11.41$ exhibit a more nearly shared linear trend; no lines are fitted in this panel.}
\label{fig:response-regions}
\end{figure}
Figure~\ref{fig:response-regions} shows different affine slopes and intercepts across depth ranges, revealing systematic depth-dependent drift that a single scale and offset cannot remove.
Region-aware Scale Adaptation~\citep{fan2025region} instead uses Segment Anything and fits a separate affine map in each region.
Under sparse or incomplete observations, many regions contain too few anchors to constrain their parameters.
Adding local affine maps therefore does not determine which global response linearizes the observed relation.

\begin{figure}[!ht]
\centering
\includegraphics[width=\textwidth]{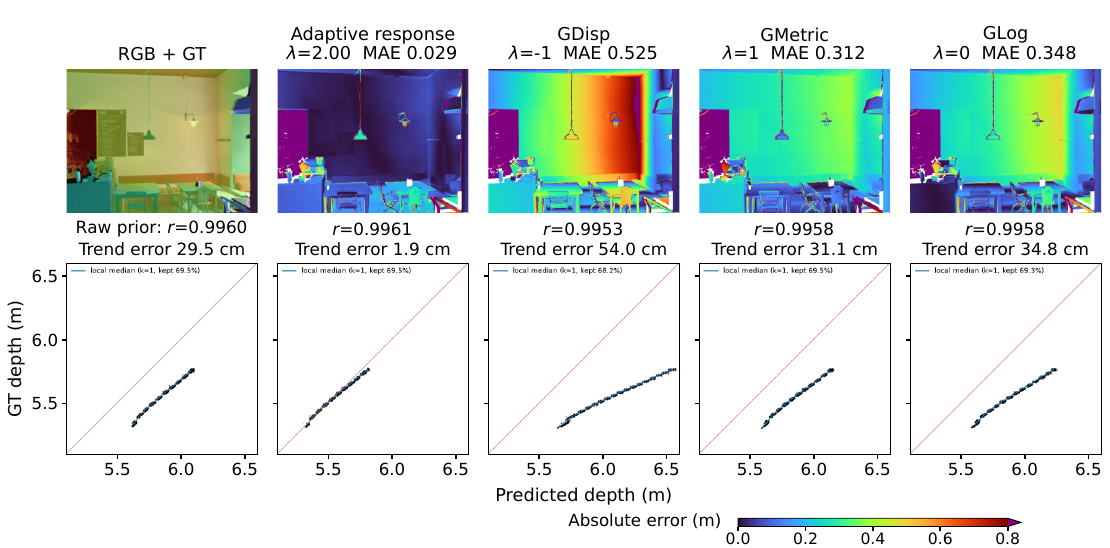}
\caption{\textbf{Depth-dependent distortion remains after an affine fit in a fixed space.}
GDisp, GMetric, and GLog denote global affine alignment in disparity, metric depth, and log-depth, respectively.
Top: scene reference and full-scene error maps.
Bottom: predicted versus ground-truth depth in meters; blue paths trace local medians, and red denotes metric identity.
The first column contains the scene reference above and the raw-prior relation below.
Pearson $r$ characterizes the displayed relation; MAE measures held-out depth error.
Trend error is the mean absolute vertical distance from the local-median path to metric identity; see Appendix~\ref{app:visualization}.}
\label{fig:motivation}
\label{fig:depth-related-distortion}
\end{figure}

A natural starting point is to examine the three coordinates commonly used by existing models: metric depth, log-depth, and disparity.
However, as Fig.~\ref{fig:motivation} shows, global affine alignment in these coordinates can still leave systematic depth-dependent deviations.
Within a fixed coordinate, one scale and offset must compromise across depth ranges requiring different corrections.
Individual ranges can therefore remain systematically overestimated or underestimated after overall alignment.
Assessing the coordinate requires examining these depth-dependent deviations, beyond matching the overall scale.

\FloatBarrier

Fig~\ref{fig:motivation} shows that the local-median relation remains curved in metric depth, log-depth, and disparity, producing structured full-scene errors.
Thus, a fixed handcrafted space need not remove the image-specific depth distortion.
On this iBims example, the estimated response is $\lambda=2.00$, outside the three conventional cases $\lambda\in\{1,0,-1\}$.

To address this limitation, we propose \emph{adaptive response geometry}, a training-free conversion from a dense relative-depth prior to metric depth.
We make the response coordinate an image-level unknown.
For a positive prior value $h$ and metric depth $s$, a monotone response $\phi_\lambda$ defines their alignment:
\begin{equation}
\phi_\lambda(s)=\alpha\phi_\lambda(h)+\beta,\qquad \alpha>0.
\label{eq:intro}
\end{equation}
$\phi_\lambda$ contains the three conventional spaces as special cases and connects them through one global parameter $\lambda$.
Changing $\lambda$ changes the relative gain assigned to near and far depth increments.
It can therefore absorb a smooth depth-dependent deformation before $\alpha$ and $\beta$ account for the remaining global scale and offset.
The response preserves depth ordering and is estimated solely from observed pairs, without access to model weights, training data, or the original loss.
The experiments use MoGe-3 as the prior, but the formulation applies to any dense positive monocular depth field.

All observed pairs constrain the shared parameters $(\lambda,\alpha,\beta)$ through squared residuals in meters, without requiring separate coefficients for unobserved objects.
We then fix these parameters and propagate the remaining log-depth residual through hard-constrained reconstruction, separating global calibration from spatial correction and enforcing every observation exactly.

We evaluate the method on DIODE, ETH3D, iBims-1, ARKitScenes, and five Tanks and Temples scenes under distributed and spatially missing observations.
Without training a completion network, our method achieves macro AbsRel $0.0301$ and median normal error $14.04^\circ$, improving both aggregate measures over PriorDA, LDCM, and Any2Full.
Ablations further isolate the contributions of adaptive response estimation and hard-constrained residual completion.

In summary, our contributions are threefold:
\begin{enumerate}
\item We formulate the response used for relative-to-metric alignment as an image-level unknown, allowing systematic depth-dependent distortion to be estimated rather than inherited from a fixed space.
\item We introduce a one-parameter monotone family that unifies metric depth, log-depth, and disparity, and jointly estimate the response and affine alignment from metric residuals.
\item We combine this global estimator with hard-constrained residual reconstruction to obtain a training-free completion method for sparse and spatially incomplete observations.
\end{enumerate}

\FloatBarrier

\section{Related Work}
\paragraph{Monocular depth prediction.}
A single image determines metric depth only up to a global scale, and often an offset.
Foundation models select the coordinate of this field during training and define the loss in that coordinate.
MiDaS~\citep{ranftl2022towards} and Depth Anything~\citep{yang2024depth} use disparity.
Their loss is invariant to a scale and a shift in inverse depth.
ZoeDepth~\citep{bhat2023zoedepth} retains a disparity backbone and supervises the metric head with a scale-invariant loss in log-depth.
Metric3D~\citep{yin2023metric3d} maps the labels to a canonical camera and applies the same scale-invariant log loss.
MoGe-3~\citep{kong2026moge3} predicts a point map and aligns metric depth with a scale and a shift along the optical axis.
However, the paired observations retain a systematic drift that varies with depth.

\paragraph{Depth completion.}
Depth completion recovers a dense metric-depth map from an image and incomplete metric observations, typically from a LiDAR scanner or an RGB-D camera.
On unobserved pixels, depth comes from a relation in the monocular prior or from a spatial regularizer~\citep{cho2026oasis,perez2003poisson}.
Gradient-domain methods fix a structural coordinate and integrate monocular guidance with the sparse anchors~\citep{zuo2024ogni,zuo2025omni,yu2026ldcm}.
Relative-to-metric methods fit a low-dimensional map from the prior to metric depth~\citep{marsal2024foundation,fan2025region,yang2026zeroshoot,ma2026midas}.
Trained completion models learn a network that takes the image, the prior, and the incomplete depth~\citep{wang2026priorda,yu2026ldcm,zhou2026any2full}.

\paragraph{Gradient-domain reconstruction.}
OGNI-DC~\citep{zuo2024ogni} reconstructs depth from sparse observations and predicted depth gradients.
OMNI-DC~\citep{zuo2025omni} integrates log-depth, while LDCM~\citep{yu2026ldcm} uses log-gradients of a monocular prior in a coarse Poisson alignment before learned refinement.
The structural penalty is shared, but the coordinate inside it is fixed in advance:
\begin{equation}
E_m(D)=E_{\mathrm{obs}}^{(m)}(D,S)+\mu_m\sum_p
\|\nabla\phi_{\lambda_m}(D)(p)-G_m(p)\|_2^2.
\label{eq:related}
\end{equation}
The observation term and the guidance $G_m$ remain specific to each method.
Raw depth and log-depth set $\lambda_m$ to $1$ and $0$, so $\nabla D$ weights increments uniformly, whereas $\nabla\log D=\nabla D/D$ weights them relative to depth.
Our completion separates this structural coordinate from the response of the global map.
Section~3.1 selects the latter.
After $(\lambda^\star,\alpha^\star,\beta^\star)$ are estimated, spatial guidance follows the gradient of $\log t$, and these parameters stay fixed.

\paragraph{Relative-to-metric mappings.}
Global disparity rescaling~\citep{marsal2024foundation} fits an affine relation in a prescribed coordinate.
When that coordinate leaves a depth-dependent residual, later methods move the remaining freedom into the map itself.
Region-aware Scale Adaptation~\citep{fan2025region} uses spatially varying affine fits, nonlinear rescaling~\citep{yang2026zeroshoot} considers richer families, and Midas Touch~\citep{ma2026midas} combines segment-level optimization with pixel refinement.
These local parameters become underconstrained when the metric support is sparse.
We therefore retain one image-global affine relation and adapt its response coordinate.
The metric mapping is nonlinear, while its freedom remains that of one response family.

\Needspace{12\baselineskip}

\section{Method}
Given a dense relative-depth prior and sparse metric observations, our goal is to recover a dense metric-depth map.
We first introduce an adaptive response family and jointly estimate its coordinate and affine alignment in metric space.
We then propagate the remaining anchor residuals through a hard-constrained spatial refinement.

\subsection{Linearity as a Coordinate Choice}
Let $I$ be an RGB image defined on a pixel grid $\Omega=\{(u,v)\}$.
Given $I$, a monocular depth foundation model provides a positive relative-depth prior $h(u,v)$ at every pixel.
Sparse or incomplete metric observations provide positive depths $s(u,v)$ on a subset $\Omega_S\subset\Omega$, typically from a LiDAR scanner or an RGB-D camera.
We collect these values as
$H=\{h(u,v)\mid(u,v)\in\Omega_S\}$ and
$S=\{s(u,v)\mid(u,v)\in\Omega_S\}$.

One common approach estimates this conversion by fitting a global affine map in a fixed response coordinate:

\begin{align}
\text{metric space:}\quad & S=\alpha H+\beta, \label{eq:fixed-metric}\\
\text{log space:}\quad &\log S=\alpha \log H+\beta,\\
\text{disparity space:}\quad & S^{-1}=\alpha H^{-1}+\beta.
\label{eq:fixed}
\end{align}

These formulations are exact only when the selected coordinate removes systematic depth-dependent differences between $H$ and $S$.
However, a coordinate fixed by the output convention of the relative-depth prior need not satisfy this condition, as illustrated in Fig.~\ref{fig:depth-related-distortion}.
The resulting affine fit leaves structured residuals across depth.
A single pair $(\alpha,\beta)$ cannot express residual structure that changes across the image.
Region-aware Scale Adaptation~\citep{fan2025region} therefore segments the scene with Segment Anything and fits a separate scale and shift in each region.
These local parameters become underconstrained when the observations are sparse or spatially incomplete.
Instead, we retain one global affine map and adapt its response coordinate using a single parameter $\lambda$:
\begin{equation}
S= \phi_{\lambda}^{-1} \big( \alpha \phi_\lambda(H)+\beta \big) ,\qquad \alpha>0.
\label{eq:affine}
\end{equation}
Here, $\phi_\lambda$ maps both relative and metric depth into a shared response coordinate, in which $\alpha$ and $\beta$ define the affine alignment.

We define this response using a scaled Box--Cox transformation~\citep{box1964analysis}.
Let $s_0>0$ be the geometric mean of the observed metric depths,
\begin{equation}
s_0=\exp\left(\frac{1}{|S|}\sum_{z\in S}\log z\right).
\label{eq:scale}
\end{equation}
For $z>0$,
\begin{equation}
\phi_\lambda(z)=
\begin{cases}
\log(z/s_0),&\lambda=0,\\[1mm]
\displaystyle\frac{(z/s_0)^\lambda-1}{\lambda},&\lambda\ne0.
\end{cases}
\label{eq:family}
\end{equation}
The logarithmic branch is the continuous limit of the power branch:
\begin{equation}
\lim_{\lambda\to 0}\frac{(z/s_0)^\lambda-1}{\lambda}=\log(z/s_0).
\label{eq:limit}
\end{equation}
For positive depth, its derivative and inverse are
\begin{equation}
\phi_\lambda'(z)=s_0^{-\lambda}z^{\lambda-1}>0,
\qquad
\phi_\lambda^{-1}(u)=
\begin{cases}
s_0\,e^{u},&\lambda=0,\\
s_0(1+\lambda u)^{1/\lambda},&\lambda\ne0,
\end{cases}
\label{eq:inverse}
\end{equation}
where $1+\lambda u>0$ when $\lambda\ne0$.
Since $\phi_\lambda'(z)>0$, the transformation is strictly increasing and therefore preserves depth order.

The parameter $\lambda$ continuously connects metric depth, log-depth, and disparity, which are otherwise treated as separate design choices.
In particular, $\lambda=1,0,-1$ gives
\begin{equation}
\phi_1(z)=\frac{z}{s_0}-1,\qquad
\phi_0(z)=\log\frac{z}{s_0},\qquad
\phi_{-1}(z)=1-\frac{s_0}{z}.
\label{eq:special}
\end{equation}
Substituting these members into Eq.~\eqref{eq:affine} recovers the three relations in Eqs.~\eqref{eq:fixed-metric}--\eqref{eq:fixed}.
Terms that depend only on $s_0$ and $\beta$ are absorbed into an offset $C$, whereas the gain remains $\alpha$:
\begin{equation}
S=\alpha H+C,\qquad
\log S=\alpha\log H+C,\qquad
S^{-1}=\alpha H^{-1}+C.
\label{eq:members}
\end{equation}

\paragraph{Geometry of depth changes.}
A systematic distortion changes the local gain between relative and metric depth as depth varies.
A fixed coordinate prescribes this gain in advance, leaving the depth-dependent component in the residual.
In contrast, $\phi_\lambda$ reweights each depth increment by a learnable power of its depth, allowing this component to be absorbed into the response coordinate before affine alignment.
For $z>0$, this reweighting is given by
\begin{equation}
d\phi_\lambda(z)=s_0^{-\lambda}z^{\lambda-1}\,dz.
\label{eq:differential}
\end{equation}
The factor $z^{\lambda-1}$ reweights an increment according to its depth.
At $\lambda=1$ the weight is constant, so equal increments of depth receive equal weight.
At $\lambda=0$, $d\phi_0(z)=dz/z$, and the coordinate records a relative change.
At $\lambda=-1$, the weight varies as $z^{-2}$ and emphasizes increments at smaller depths.

The same response controls the spatial gradient of the relative-depth field:
\begin{equation}
\nabla\phi_\lambda(h)=s_0^{-\lambda}h^{\lambda-1}\nabla h.
\label{eq:spatial}
\end{equation}
For a paired depth relation, differentiating
$\phi_\lambda(S)=\alpha\phi_\lambda(H)+\beta$ gives
\begin{equation}
S^{\lambda-1}\,dS=\alpha\, H^{\lambda-1}\,dH,
\qquad
\frac{d\log S}{d\log H}=\alpha\left(\frac{H}{S}\right)^\lambda.
\label{eq:gain}
\end{equation}
Although $\alpha$ is shared across the image, the metric-depth slope $dS/dH=\alpha(H/S)^{\lambda-1}$ can vary with depth.
The offset $\beta$ does not affect this local gain.
This depth-dependent gain accommodates systematic calibration drift of the kind illustrated in Fig.~\ref{fig:response-regions}, where different depth ranges exhibit different local affine relations.
Estimating $\lambda$ lets a single global response account for these differences without introducing separate affine parameters for each region.

\subsection{Metric-Space Estimation}
We select the response coordinate by its metric-depth error.
We therefore estimate $(\lambda,\alpha,\beta)$ directly from the paired values in $H$ and $S$.
For the $n=|\Omega_S|$ observed pairs $(H_i,S_i)$, Eq.~\eqref{eq:affine} gives the metric prediction
\begin{equation}
t_i(\lambda,\alpha,\beta)
=\phi_\lambda^{-1}\big(\alpha\phi_\lambda(H_i)+\beta\big).
\label{eq:prediction}
\end{equation}
We minimize the residuals in meters:
\begin{equation}
(\lambda^\star,\alpha^\star,\beta^\star)
=\arg\min_{\alpha>0}
\sum_{i=1}^{n}\big(S_i-t_i(\lambda,\alpha,\beta)\big)^2.
\label{eq:estimator}
\end{equation}
For $\lambda\neq0$, this prediction is defined only if
\begin{equation}
1+\lambda\big(\alpha\phi_\lambda(H_i)+\beta\big)>0
\qquad\text{for every }i.
\label{eq:feasibility}
\end{equation}

This objective is nonlinear in $\lambda$.
To solve it while enforcing $\alpha>0$, we parameterize
$\theta=(\lambda,\log\alpha,\beta)$ and define
$r_i(\theta)=S_i-t_i$.
Let $J(\theta)\in\R^{n\times3}$ denote the Jacobian of
$r=(r_1,\ldots,r_n)$ with respect to $\theta$.
Using the implicit relation
$\phi_\lambda(t_i)=\alpha\phi_\lambda(H_i)+\beta$,
its three entries are
\begin{equation}
\frac{\partial r_i}{\partial\lambda}
=\frac{\dot{\phi}_\lambda(t_i)-\alpha\dot{\phi}_\lambda(H_i)}
{\phi_\lambda'(t_i)},\qquad
\frac{\partial r_i}{\partial\log\alpha}
=-\frac{\alpha\phi_\lambda(H_i)}{\phi_\lambda'(t_i)},\qquad
\frac{\partial r_i}{\partial\beta}
=-\frac{1}{\phi_\lambda'(t_i)},
\label{eq:jacobian}
\end{equation}
where $\dot{\phi}_\lambda(z)=\partial\phi_\lambda(z)/\partial\lambda$.
For $\lambda\ne0$, this derivative is
\begin{equation}
\dot{\phi}_\lambda(z)
=\frac{\lambda(z/s_0)^\lambda\log(z/s_0)
-\big((z/s_0)^\lambda-1\big)}{\lambda^2},
\qquad
\dot{\phi}_0(z)=\frac{1}{2}\log^2(z/s_0).
\label{eq:lambda-derivative}
\end{equation}
At each iteration, Levenberg--Marquardt~\citep{marquardt1963algorithm} solves
\begin{equation}
\big(J^\top J+\mu I\big)\delta=-J^\top r,
\qquad
\theta\leftarrow\theta+\delta.
\label{eq:lm}
\end{equation}
Since $\theta$ contains only three variables, the system matrix in Eq.~\eqref{eq:lm} is $3\times3$.
We accept a step when it reduces $\|r\|_2^2$ and satisfies Eq.~\eqref{eq:feasibility}, after which we decrease the damping coefficient $\mu$.
Otherwise, we reject the step and increase $\mu$.

\subsection{Hard-Constrained Completion}
The estimated parameters define a global conversion over the image.
However, the least-squares fit does not guarantee exact agreement with every metric observation.
We therefore separate global metric alignment from spatial residual correction.
Specifically, we first apply the estimated conversion to the complete relative-depth field:
\begin{equation}
t(u,v)
=\phi_{\lambda^\star}^{-1}\big(
\alpha^\star\phi_{\lambda^\star}(h(u,v))+\beta^\star
\big).
\label{eq:dense}
\end{equation}
We then propagate the remaining residuals by Poisson reconstruction~\citep{perez2003poisson}.
Following OASIS-DC~\citep{cho2026oasis}, we impose the metric observations as Dirichlet constraints and use the gradient of $\log t$ for spatial guidance.
Gradient matching alone preserves local structure but leaves the solution weakly anchored away from the observations.
We therefore regularize both the magnitude and curvature of the log residual.
Let $x=\log D$ denote the refined log-depth map and $\ell=\log t$ denote the globally aligned prediction.
We solve
\begin{align}
\min_x\quad&\lambda_{\rm grad}\|\nabla x-\nabla \ell\|_2^2
+\lambda_{\rm data}\|x-\ell\|_2^2
+\lambda_{\rm lap}\|\Delta x-\Delta \ell\|_2^2,\nonumber\\
\text{subject to}\quad&x(u,v)=\log s(u,v),\qquad (u,v)\in\Omega_S,
\label{eq:completion}
\end{align}
with $(\lambda_{\rm grad},\lambda_{\rm data},\lambda_{\rm lap})=(1,10^{-3},10^{-3})$.
The data term restricts the magnitude of the propagated correction, whereas the Laplacian term suppresses curvature unsupported by $\ell$.
The Dirichlet constraints enforce $D(u,v)=s(u,v)$ on $\Omega_S$ exactly.
This spatial refinement corrects the residual field without changing the estimated parameters $(\lambda^\star,\alpha^\star,\beta^\star)$.

\Needspace{12\baselineskip}
\section{Experiments}

\subsection{Incomplete Metric Support}
The evaluation uses indoor images from DIODE~\citep{vasiljevic2019diode}, ETH3D~\citep{schops2017mvs}, iBims-1~\citep{koch2018evaluation}, and ARKitScenes~\citep{baruch2021arkitscenes}, together with five Tanks and Temples scenes~\citep{knapitsch2017tanks}: Barn, Caterpillar, Ignatius, Meetingroom, and Truck.
Median valid reference coverage ranges from $0.1554$ on Ignatius to $0.9938$ on DIODE-in.

The evaluation combines distributed sparse observations with central and outer removal on incomplete reference support; protocol and coverage details appear in Appendix~\ref{app:experimental-details}.
Evaluation excludes observed pixels.

Every variant uses the MoGe-3 prior~\citep{kong2026moge3}.
We compare the complete method with PriorDA~\citep{wang2026priorda}, LDCM~\citep{yu2026ldcm}, Any2Full~\citep{zhou2026any2full}, and PromptDA~\citep{lin2025promptda}.
Ours w/o Dirichlet denotes response estimation alone; Ours includes hard-constrained residual completion.
Figure~\ref{fig:joint} also includes the raw MoGe-3 prior, fixed-coordinate alignment, nearest-anchor regression (KNN), locally weighted linear regression (LWLR), and an LDCM-style log-Poisson baseline~\citep{yu2026ldcm}; Appendix~\ref{app:baselines} specifies these configurations.
We report absolute relative error (AbsRel) and the per-image median angular normal error (NMed).
We report median valid per-image scores (p50) for all nine subsets separately and their unweighted mean as Macro; Appendix~\ref{app:percentiles} extends the comparison to p7--p93.

\subsection{Response Coordinate and Metric Alignment}
We examine response estimation before spatial completion.
Figure~\ref{fig:motivation} connects the response coordinate to metric alignment on one iBims image.
The estimated response is $\lambda=2.00$, outside the fixed members $\lambda\in\{1,0,-1\}$.
Its local-median path lies closer to metric identity than the paths of those three coordinates, and the scene error changes with the coordinate.

\begin{figure}[!ht]
\centering
\setlength{\abovecaptionskip}{2pt}
\includegraphics[width=\textwidth]{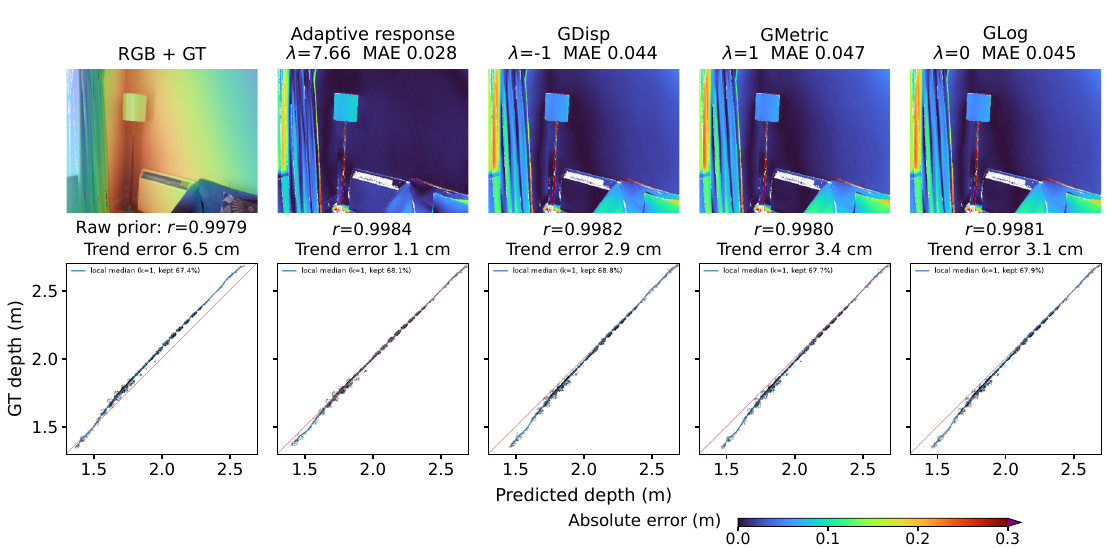}
\caption{\textbf{A near-linear relation can still leave metric error.}
The first column pairs the scene reference with the raw-prior relation; each remaining column pairs one response with its relation.}
\label{fig:arkit}
\end{figure}

Figure~\ref{fig:arkit} examines the same choice when the fixed coordinates are already highly correlated with metric depth.
Adaptive estimation selects $\lambda=7.66$.
The displayed Pearson correlation is $0.9984$, compared with $0.9980$--$0.9982$ for $\lambda\in\{1,0,-1\}$, and held-out MAE falls from $0.044$--$0.047$\,m to $0.028$\,m.
Near $r=1$, this gap is small on the scale of $r$ and larger relative to the remaining departure $1-r$.
The MAE and the error maps show that the change remains visible in metric depth.

Appendix~\ref{app:visualization} explains the diagnostic measures and display conventions.

\FloatBarrier
\subsection{Comparison with Trained Completion Methods}
\begin{table}[H]
\centering
\setlength{\abovecaptionskip}{2pt}
\setlength{\belowcaptionskip}{2pt}
\caption{\textbf{Per-subset comparison of complete methods.} Lower is better for AbsRel and NMed.}
\label{tab:main}
\small
\setlength{\tabcolsep}{3pt}
\renewcommand{\arraystretch}{0.92}
\begin{tabular*}{\textwidth}{@{\hspace{2pt}\extracolsep{\fill}}lr@{\hspace{5.25pt}}rr@{\hspace{5.25pt}}rr@{\hspace{5.25pt}}rr@{\hspace{5.25pt}}rr@{\hspace{5.25pt}}r@{\hspace{2pt}}}
\toprule
& \multicolumn{2}{c}{PriorDA} & \multicolumn{2}{c}{LDCM} & \multicolumn{2}{c}{Any2Full} & \multicolumn{2}{c}{PromptDA$^{\ast}$} & \multicolumn{2}{c}{Ours}\\
\cmidrule(lr){2-3}\cmidrule(lr){4-5}\cmidrule(lr){6-7}\cmidrule(lr){8-9}\cmidrule(lr){10-11}
Subset & AbsRel & NMed & AbsRel & NMed & AbsRel & NMed & AbsRel & NMed & AbsRel & NMed\\
\midrule
DIODE-in & 0.0300 & 19.63 & 0.0281 & 13.04 & 0.0307 & 12.30 & 0.9943 & 77.59 & \textbf{0.0275} & \textbf{10.05}\\
ETH3D & 0.0228 & 24.34 & 0.0187 & 14.03 & 0.0254 & 13.12 & 0.9939 & 60.90 & \textbf{0.0180} & \textbf{11.52}\\
iBims & 0.0250 & 18.87 & \textbf{0.0203} & 11.67 & 0.0229 & 11.26 & 0.9939 & 65.96 & 0.0208 & \textbf{10.83}\\
ARKit & 0.0163 & 12.74 & \textbf{0.0145} & 9.40 & 0.0169 & 10.05 & 0.9876 & 71.84 & 0.0154 & \textbf{8.90}\\
\midrule
Barn & 0.0276 & 16.37 & \textbf{0.0207} & 11.19 & 0.0362 & 11.85 & 0.9944 & 64.29 & 0.0251 & \textbf{10.43}\\
Caterpillar & 0.0499 & 31.69 & 0.0792 & 31.38 & 0.0571 & \textbf{21.10} & 0.9960 & 74.32 & \textbf{0.0469} & 22.65\\
Ignatius & 0.0273 & 27.80 & 0.0547 & 29.13 & 0.0326 & 24.60 & 0.9983 & 73.05 & \textbf{0.0197} & \textbf{18.05}\\
Meetingroom & 0.0631 & 34.14 & \textbf{0.0507} & 23.33 & 0.0726 & 19.98 & 0.9917 & 66.68 & 0.0569 & \textbf{19.38}\\
Truck & 0.0417 & 22.14 & 0.0446 & 15.36 & 0.0513 & \textbf{14.09} & 0.9933 & 61.97 & \textbf{0.0404} & 14.55\\
\midrule
Macro & 0.0337 & 23.08 & 0.0368 & 17.61 & 0.0384 & 15.37 & 0.9937 & 68.51 & \textbf{0.0301} & \textbf{14.04}\\
\bottomrule
\end{tabular*}

\par\vspace{2pt}
{\scriptsize Five subsets from Tanks and Temples; see Appendix~\ref{app:tt-selection}.\quad $^{\ast}$PromptDA is analyzed in Appendix~\ref{app:promptda}.\par}
\end{table}
Table~\ref{tab:main} compares Ours with four trained systems.
Our method attains macro AbsRel $0.0301$ and NMed $14.04^\circ$.
Both aggregates are lower than those of PriorDA ($0.0337$, $23.08^\circ$), LDCM ($0.0368$, $17.61^\circ$), and Any2Full ($0.0384$, $15.37^\circ$).
Relative to the strongest competing aggregate, AbsRel is $0.0036$ below PriorDA and NMed is $1.33^\circ$ below Any2Full.

\begin{figure}[H]
\centering
\setlength{\abovecaptionskip}{2pt}
{\raggedright\small\textbf{(a) Macro comparison}\par}
\includegraphics[width=\textwidth,trim=0 5pt 0 0,clip]{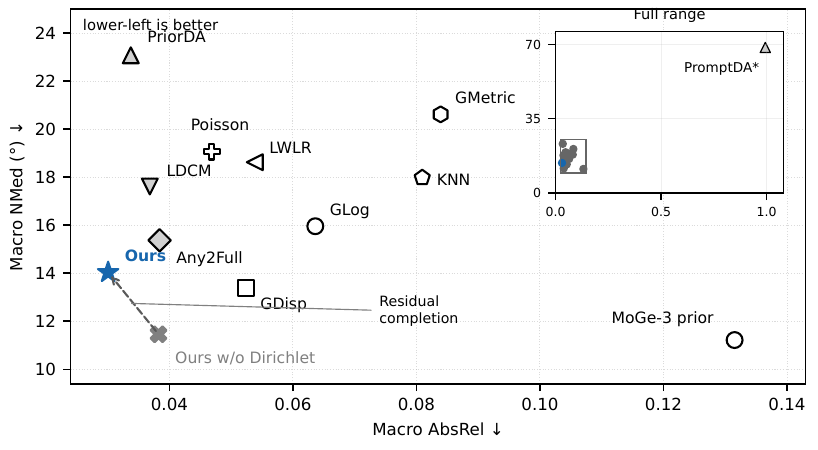}
\par\vspace{2pt}
{\raggedright\small\textbf{(b) Per-subset comparison at p50}\par}
\includegraphics[width=\textwidth]{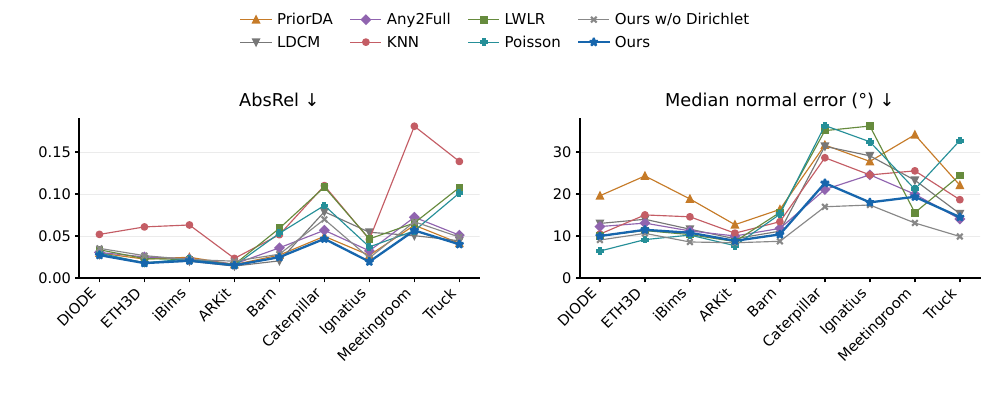}
\caption{\textbf{Metric depth and surface geometry.}
(a) Macro comparison; the inset includes PromptDA and boxes the main-axis range. The arrow links Ours w/o Dirichlet to Ours.
(b) Per-subset p50 results: depth error on the left and normal error on the right. The two views show aggregate performance and its variation across subsets.
Baseline settings and stage normal quality are detailed in Appendices~\ref{app:baselines} and~\ref{app:support}.}
\label{fig:joint}
\end{figure}

Among complete systems, Ours has the lowest AbsRel on five subsets and NMed on seven.
LDCM has lower AbsRel on iBims, ARKit, Barn, and Meetingroom; Any2Full has lower NMed on Caterpillar and Truck.
Appendix~\ref{app:dataset-comparison} details the per-dataset comparison.
Figure~\ref{fig:joint} places the complete method beside the fixed coordinates, the raw prior, and Poisson reconstruction.
Appendix~\ref{app:percentiles} reports the distributions beyond p50.

\begin{figure}[H]
\centering
\includegraphics[width=\textwidth]{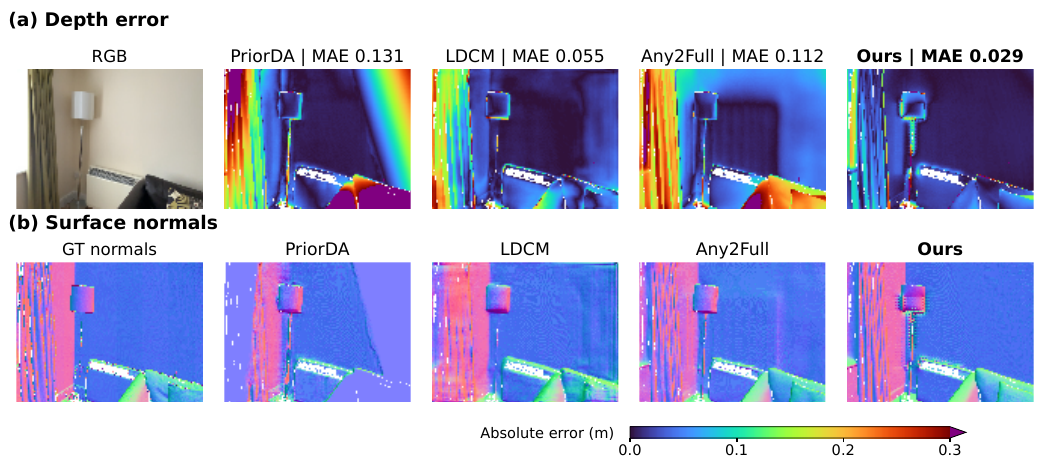}
\caption{\textbf{ARKit full-scene comparison.}
Columns show the reference, PriorDA, LDCM, Any2Full, and our method.
(a) RGB reference and absolute depth-error maps; each title reports held-out MAE in meters.
(b) Ground-truth normals and predicted normals for the same image.
All error maps share the displayed scale, with purple above $0.3$\,m.}
\label{fig:qualitative}
\end{figure}

Figure~\ref{fig:qualitative} compares the four complete systems on one ARKit image.
Held-out MAE is $0.029$\,m for our method, compared with $0.131$\,m for PriorDA, $0.055$\,m for LDCM, and $0.112$\,m for Any2Full.

On the right wall, PriorDA shifts orientation over a broad region, while LDCM and Any2Full vary more strongly across the surface.
Our reconstruction stays closer to the reference wall orientation and has lower depth error over much of that wall and the foreground cushion.
The lamp remains a localized source of depth and normal error.
The paired maps show that low pointwise depth error and consistent surface orientation are distinct aspects of the reconstruction: the wall can be assessed over a broad surface, while the lamp exposes a remaining local limitation.

\FloatBarrier

\subsection{Response Estimation and Residual Completion}
Ours w/o Dirichlet obtains macro AbsRel $0.0382$ and NMed $11.46^\circ$, compared with $0.0524$ and $13.38^\circ$ for fixed disparity alignment.
Selecting $\lambda$ therefore lowers both aggregates while using only the global parameters $(\lambda,\alpha,\beta)$, before spatial correction is introduced.
The response stage changes the global depth relation while retaining lower normal error.
Dirichlet completion further reduces macro AbsRel to $0.0301$, with an improvement on every subset, while NMed increases to $14.04^\circ$.
Spatial correction improves agreement with metric observations, while the resulting changes in local depth gradients can increase surface-normal error.
The residual stage enforces the observations and improves depth accuracy, but changes local surface orientation as the residual is propagated.

The complete method nevertheless has lower macro NMed than all four trained baselines, together with lower macro AbsRel.
These measurements separate the two stages of the method from the complete-system comparison in Table~\ref{tab:main}.
Figure~\ref{fig:joint} shows both stages, and Appendix~\ref{app:additional} examines their normal quality and reconstruction weights in detail.

\FloatBarrier
\Needspace{10\baselineskip}

\section{Discussion and Conclusion}
Adaptive response geometry makes the coordinate of affine depth alignment an estimated quantity.
Its exponent specifies a depth-dependent gain while the affine parameters remain global, allowing alignment to vary with depth without fitting a separate map to each image region.
This distinction matters under incomplete support: an unobserved region still receives the same dense conversion estimated from the available pairs.
Hard-constrained completion then incorporates spatial corrections without refitting that global relation.
The resulting training-free method improves both macro depth accuracy and median normal error over the four trained comparison systems.

The formulation also separates two limits of reconstruction.
A single monotone response cannot resolve arbitrary spatial disagreements or correct depth-order reversals in the prior; these are outside the freedom provided by the exponent.
The hard constraints use reference measurements from the geometry-screened benchmark described in Appendix~\ref{app:tt-selection}.
The present results establish the benefit of the combined pipeline, while the stage comparison identifies geometry already present before residual reconstruction.
Appendix~\ref{app:additional} develops this distinction through the normal results and weight sensitivity, complementing the complete-system comparisons in the main text.

\section*{Reproducibility Statement}
The accompanying source includes the result CSVs, the full-scene visual assets, and a script that plots the recorded percentile entries directly. Numerical optimization and completion settings are specified in Appendix~\ref{app:implementation}; diagnostic visualization conventions appear in Appendix~\ref{app:visualization}.

\section*{AI use statement}
We used GPT-5.6 Sol \citep{openai2026gpt56sol} to assist with retrieving
and organizing related work. We used GPT-6 Astra
\citep{openai2026gpt6astra} to improve the conciseness of portions of the
manuscript, organize transitions, and occasionally assist with translation.
In addition, some figures in the paper were generated using Claude Code
\citep{anthropic2026claudecode} based on experimental data produced by the
authors.
We reviewed and verified all AI-assisted outputs and take responsibility for the final content of this work.

\clearpage
\appendix
\newlength{\appTableWidth}
\renewcommand{\thetable}{\Alph{section}.\arabic{table}}
\makeatletter
\@addtoreset{table}{section}
\setlength{\@fptop}{0pt}
\makeatother
\clearpage
\section{Adaptive Response Geometry: Further Details}
\label{app:section-A}
\label{app:implementation}

\subsection{Numerical Evaluation and Dense-Domain Feasibility}
For the response defined in Eq.~\eqref{eq:family}, let $q=\log(z/s_0)$.
The nonzero-$\lambda$ branch can be evaluated as $\operatorname{expm1}(\lambda q)/\lambda$, avoiding subtraction of nearly equal numbers when $\lambda q$ is small.
The implementation switches to the logarithmic branch for $|\lambda|<10^{-5}$.

The inverse-domain condition is checked over the full valid range of the dense prior.
Let $h_{\min}$ and $h_{\max}$ be its positive valid extrema and define
\[
g(h)=1+\lambda\big(\alpha\phi_\lambda(h)+\beta\big).
\]
Since $\alpha>0$ and $\phi_\lambda$ is increasing, $g$ is increasing for $\lambda>0$ and decreasing for $\lambda<0$.
Consequently, feasibility over the entire prior range can be checked at $h_{\min}$ for positive $\lambda$ and at $h_{\max}$ for negative $\lambda$.
The implementation checks the endpoint domain values before accepting a dense mapping.
The logarithmic member has no corresponding finite response-domain boundary.

\subsection{Masked Residual System and Linear Solver}
Writing $r=\log D-\log t$ converts Eq.~\eqref{eq:completion} into
\[
\min_r r^\top Qr,\qquad
Q=\lambda_{\rm grad}G^\top G+\lambda_{\rm data}I+
\lambda_{\rm lap}L^\top L.
\]
Here $G$ contains horizontal and vertical first differences only between two valid pixels.
The Laplacian uses the same four-neighbor graph: $(Lr)_p=\sum_{q\sim p}(r_q-r_p)$.
Invalid pixels are excluded from both the unknown vector and its graph edges; an image or validity boundary therefore has fewer incident edges with graph connections restricted to valid neighbors.

At anchor pixels $\mathcal F$, $r_{\mathcal F}=\log s_{\mathcal F}-\log t_{\mathcal F}$ is fixed.
With $U$ denoting the free valid pixels, the system is
\[
Q_{UU}r_U=-Q_{U\mathcal F}r_{\mathcal F}.
\]
The solver applies the graph operators without explicitly assembling the full matrix.
For a valid pixel with degree $d_p$, the Jacobi diagonal is
\[
Q_{pp}=\lambda_{\rm data}+\lambda_{\rm grad}d_p+
\lambda_{\rm lap}(d_p^2+d_p).
\]
The free entries of this diagonal precondition conjugate gradients.
The stopping threshold is $10^{-6}$ for the relative linear-system residual, with at most $500$ iterations.
The implementation normalizes depths by the anchor geometric mean before taking logarithms; the shared constant cancels from $r$ and is restored when depth is recovered.
It initializes free log-depth values from the calibrated prediction and writes the original observed depths back at anchor locations after reconstruction.

Because $\lambda_{\rm data}>0$, $Q_{UU}$ is positive definite even when the valid-pixel graph has disconnected components.
An entirely unobserved component has zero anchor forcing, so its minimizing residual is zero and its prediction remains $t$.
This specifies the behavior where observation residuals cannot propagate through valid graph edges.

\subsection{Observation and scoring details}
\label{app:experimental-details}

Observations are formed on a regular low-resolution grid with nominal stride 7.5. Positive reference depths in each cell are averaged, and the resulting values are placed at the corresponding anchor locations. The same construction is used for all methods.

The central rectangle occupies half the image width and half its height. The distributed case retains the grid observations and scores non-anchor valid pixels. Central removal retains observations outside the rectangle and scores its interior; outer removal retains observations inside the rectangle and scores its exterior. These remove observations from 25\% and 75\% of the image area, respectively, acting on already incomplete reference support. The three cases are pooled into one operating distribution. Reference-support statistics and benchmark inclusion are detailed in Appendix~\ref{app:tt-selection}.

For each method, MAE and AbsRel are computed over valid scored pixels excluding anchors:

\[
\operatorname{MAE}=|M|^{-1}\sum_{p\in M}|\hat D_p-D_p|,
\qquad
\operatorname{AbsRel}=|M|^{-1}\sum_{p\in M}|\hat D_p-D_p|/D_p.
\]

Normals are obtained by back-projecting depth with camera intrinsics, taking centered horizontal and vertical differences of the 3D point map, and normalizing their cross product. The same operator is used for predictions and reference depths. Scoring uses the intersection of the returned validity masks and the evaluation region, excluding anchors and the outer pixel ring. Per-image mean and median angular errors are recorded separately.

The seven reported percentiles pool valid per-image scores across the three observation cases within each subset. Thus p50 NMed is a median across per-image median angular errors; it is not a pooled-pixel angular median. Macro is the unweighted mean of the nine subset p50 values. Missing entries and failure counts retain the conventions of the result export.

\subsection{Linearity diagnostics}
\label{app:visualization}

Figures 2 and 3 retain full-scene visual maps. Their scatter diagnostics sample up to 6,000 held-out pixels with seed 0 and use a shared displayed depth window. Predicted-depth bins are 0.05 m wide, with a minimum of 30 samples. Within-bin median/MAD filtering selects displayed points; local-median paths follow the supported bins, with gaps left where there is insufficient support. Pearson correlation is calculated on the displayed points.

Trend error is the mean absolute vertical distance of the supported local-median path from metric identity, with equal weighting over supported bins, expressed in centimeters. Pearson describes linear association; metric identity additionally fixes scale and offset. The scatter statistics and full held-out MAE have different supports.

GDisp fits an affine map by least squares in disparity; GMetric and GLog fit in metric depth and log-depth, respectively. Each fitted map is inverted to recover metric depth. Adaptive panels use the recorded response parameters. Shared depth-error and normal-orientation scales are used within each qualitative comparison.

Figure 1 uses all valid paired pixels from ETH3D office/DSC\_0223. Its colors start from GT depth-rank groups P0--P1, P1--P50, and P50--P100. RANSAC reference lines use a 0.01 m vertical-residual threshold. Blue and red groups are expanded using local residual-density support, with the original depth ranges extended by 0.10 m at either end; remaining points form the yellow group. A small blue patch enclosed by the yellow band was manually reassigned to yellow for this diagnostic visualization. No points are removed from the final figure, and these assignments are not part of the completion method.

The RGB overlay uses 50\% opacity. The middle panel retains the reference lines used during group construction; the right panel transforms the same point pairs with recorded \ensuremath{\lambda}=\ensuremath{-}11.4148 and contains no fitted lines. Prior coordinates are normalized for display after the relevant transformation. The colors identify the same pixels across panels, while the axes represent different coordinates.
\FloatBarrier

\subsection{Complete Baseline Comparisons}
\label{app:section-F}
\label{app:baselines}
\label{app:dataset-comparison}
\label{app:percentiles}

\paragraph{Methods and configurations.}

The complete-system comparison includes PriorDA, LDCM, Any2Full, PromptDA, and Ours. The raw MoGe-3 prior and three additional reconstruction baselines provide context for spatial and global correction. Table~\ref{tab:app-F-1} reports p50 AbsRel and mean normal error; Appendix~\ref{app:section-C} examines normal quality in detail.

Among the four trained baselines and Ours, the complete method has the lowest AbsRel on DIODE, ETH3D, Caterpillar, Ignatius, and Truck. LDCM is lower on iBims (0.0203 versus 0.0208), ARKit (0.0145 versus 0.0154), Barn (0.0207 versus 0.0251), and Meetingroom (0.0507 versus 0.0569). These are subset-specific comparisons; the main macro result averages all nine p50 values equally.

The raw-prior entry applies no response calibration or spatial completion. KNN fits a local affine disparity map using the five nearest observed image locations with inverse-distance weights and restores observed depths at anchors. LWLR fits affine depth maps on a 24\ensuremath{\times}24 spatial grid using Gaussian weights with standard deviation 0.1 times the larger image dimension, then interpolates the coefficients bilinearly.

Poisson denotes the benchmark's LDCM-style multiscale log-depth reconstruction baseline. It first aligns the raw prior in metric depth and integrates clipped log-gradients with a soft observation penalty. Its configuration has four pyramid levels, gradient clipping at \ensuremath{\pm}3, observation weight 5, and conjugate-gradient relative tolerance 10\ensuremath{^{-5}}. The default solve stops at quarter resolution before upsampling. Its guidance comes from the globally aligned raw prior, with soft observation constraints.

\paragraph{Performance beyond p50.}

Figure~\ref{fig:percentiles} at the end of Appendix~\ref{app:section-A} presents the detailed percentile comparison for eight methods, including KNN, LWLR, Poisson, and Ours w/o Dirichlet. PromptDA and the raw prior remain in the p50 table and CSVs; PromptDA is discussed in Appendix~\ref{app:promptda}.

The ranking varies across the distribution. On Caterpillar, Ours improves p50 AbsRel over Any2Full (0.0469 versus 0.0571), whereas Any2Full is lower at p93 (0.1069 versus 0.1174). On Ignatius, Ours remains substantially below LDCM at p93 (0.0399 versus 0.5687). Meetingroom has a different ordering, with LDCM lower than Ours at p50 and p93. These cases show how the ordering varies across scenes and error percentiles.

The table is restricted to p50; see Fig.~\ref{fig:percentiles} for the distribution comparison and the accompanying CSVs for all numerical percentiles and failure counts.

\begin{table}[H]
\centering
\caption{P50 comparison across nine subsets. NMean is the per-image mean normal error in degrees. Detailed percentile behavior is shown in Fig.~\ref{fig:percentiles}.}\label{tab:app-F-1}
\small\setlength{\tabcolsep}{3pt}
\begin{tabular*}{\textwidth}{@{\extracolsep{\fill}}lrrrrrrrrr@{}}
\toprule
\multicolumn{10}{l}{\textbf{AbsRel $\downarrow$}}\\
Method & DIODE & ETH3D & iBims & ARKit & Barn & Cat. & Ign. & Meet. & Truck\\
\midrule
Raw MoGe-3 & 0.2707 & 0.0805 & 0.0773 & 0.0678 & 0.0857 & 0.1393 & 0.2643 & 0.1055 & 0.0925\\
PriorDA & 0.0300 & 0.0228 & 0.0250 & 0.0163 & 0.0276 & 0.0499 & 0.0273 & 0.0631 & 0.0417\\
LDCM & 0.0281 & 0.0187 & 0.0203 & 0.0145 & 0.0207 & 0.0792 & 0.0547 & 0.0507 & 0.0446\\
Any2Full & 0.0307 & 0.0254 & 0.0229 & 0.0169 & 0.0362 & 0.0571 & 0.0326 & 0.0726 & 0.0513\\
PromptDA & 0.9943 & 0.9939 & 0.9939 & 0.9876 & 0.9944 & 0.9960 & 0.9983 & 0.9917 & 0.9933\\
KNN & 0.0522 & 0.0611 & 0.0634 & 0.0236 & 0.0518 & 0.1102 & 0.0462 & 0.1809 & 0.1390\\
LWLR & 0.0339 & 0.0234 & 0.0222 & 0.0164 & 0.0594 & 0.1085 & 0.0468 & 0.0658 & 0.1080\\
Poisson & 0.0299 & 0.0183 & 0.0231 & 0.0150 & 0.0535 & 0.0866 & 0.0373 & 0.0565 & 0.1014\\
Ours & 0.0275 & 0.0180 & 0.0208 & 0.0154 & 0.0251 & 0.0469 & 0.0197 & 0.0569 & 0.0404\\
\midrule
\multicolumn{10}{l}{\textbf{NMean $\downarrow$}}\\
Method & DIODE & ETH3D & iBims & ARKit & Barn & Cat. & Ign. & Meet. & Truck\\
\midrule
Raw MoGe-3 & 18.37 & 21.70 & 16.19 & 12.97 & 15.35 & 27.52 & 21.93 & 24.43 & 18.96\\
PriorDA & 29.74 & 37.69 & 29.53 & 20.32 & 25.86 & 41.48 & 33.52 & 42.49 & 33.13\\
LDCM & 24.46 & 27.07 & 22.26 & 16.36 & 19.04 & 41.17 & 36.63 & 33.88 & 26.59\\
Any2Full & 20.57 & 24.93 & 19.60 & 15.76 & 20.45 & 32.02 & 29.05 & 30.58 & 24.26\\
PromptDA & 77.26 & 67.47 & 67.68 & 73.51 & 65.62 & 74.94 & 75.28 & 69.08 & 66.56\\
KNN & 24.77 & 32.56 & 29.08 & 20.36 & 26.12 & 41.38 & 33.88 & 38.78 & 34.36\\
LWLR & 18.63 & 23.07 & 17.54 & 13.96 & 22.14 & 42.68 & 39.31 & 26.25 & 32.90\\
Poisson & 18.65 & 22.71 & 21.49 & 14.68 & 25.25 & 44.45 & 37.00 & 32.54 & 40.06\\
Ours & 20.55 & 23.84 & 20.99 & 15.07 & 19.48 & 34.37 & 24.71 & 31.11 & 27.07\\
\bottomrule
\end{tabular*}
\end{table}
\FloatBarrier

\FloatBarrier

\clearpage
\begin{figure}[p]
\centering
\includegraphics[width=\textwidth,height=.91\textheight,keepaspectratio]{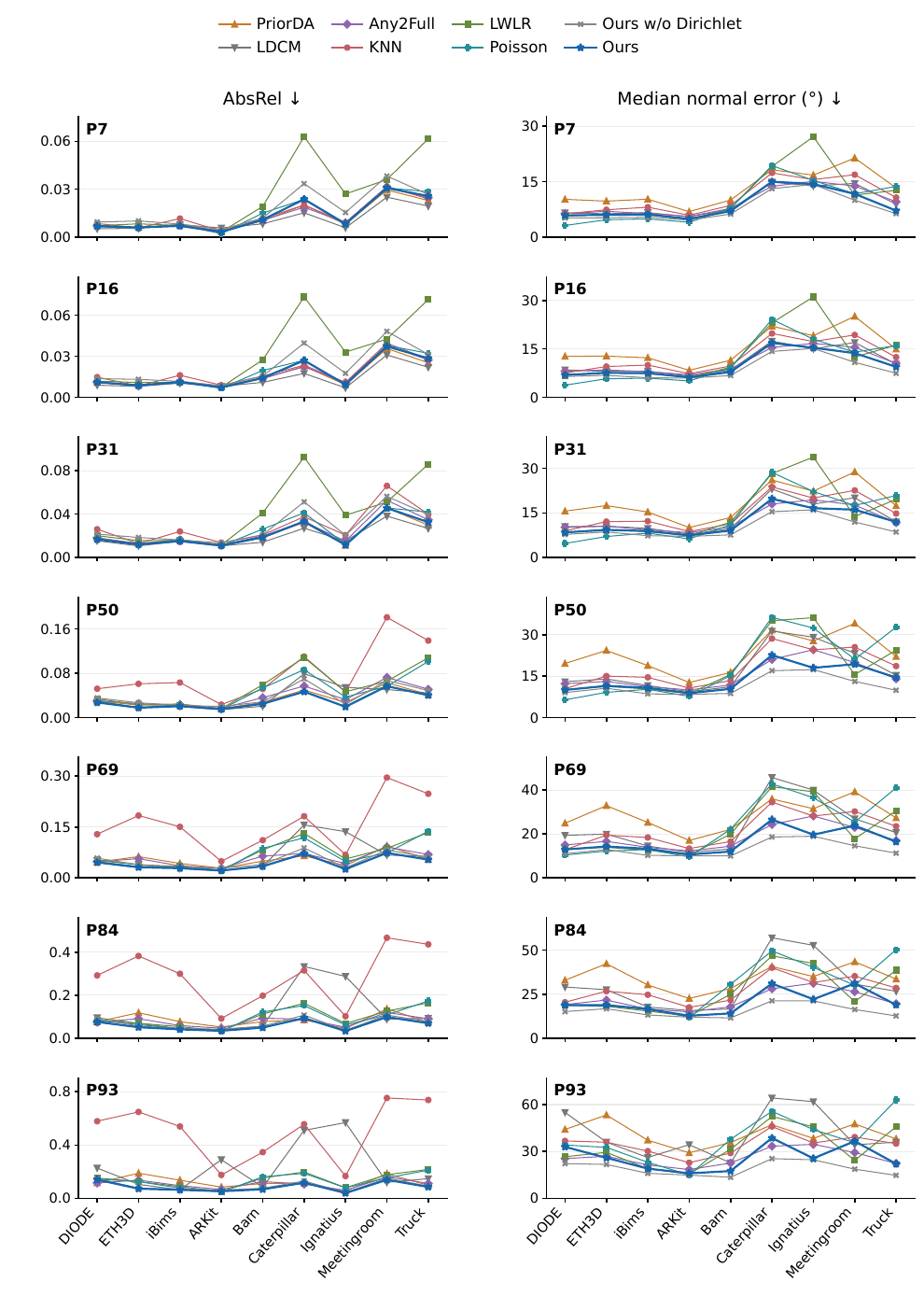}
\caption{Complete-system error distributions across nine subsets. Rows show p7--p93; columns show AbsRel and median normal error. Eight methods are shown: PriorDA, LDCM, Any2Full, KNN, LWLR, Poisson, Ours w/o Dirichlet, and Ours.}
\label{fig:percentiles}
\end{figure}
\clearpage

\clearpage
\section{Response Estimation and Residual Completion}
\label{app:section-B}
\label{app:stages-depth}

The response and residual stages contribute different forms of correction. Response estimation chooses one dense depth relation from the available metric pairs. Residual completion then enforces those pairs spatially. Table~\ref{tab:app-B-1} reports p50 depth errors for both stages across all nine subsets.

Dirichlet completion reduces AbsRel on every subset. Caterpillar changes from 0.0701 to 0.0469, while ETH3D changes from 0.0269 to 0.0180. MAE decreases on eight subsets; iBims changes slightly from 0.0604 m to 0.0610 m despite its lower AbsRel. The two depth metrics weight errors differently.

The stage scatter summarizes these paired changes. Each point compares response-only and complete p50 AbsRel for the same subset; all nine lie below the equality line.

The p50 table directly accompanies the stage plot. Both stages use the same monocular prior, isolating the effect of the residual solve. Surface orientation is examined separately in Appendix~\ref{app:section-C}.

\begin{figure}[H]
\centering
\includegraphics[width=.75\textwidth]{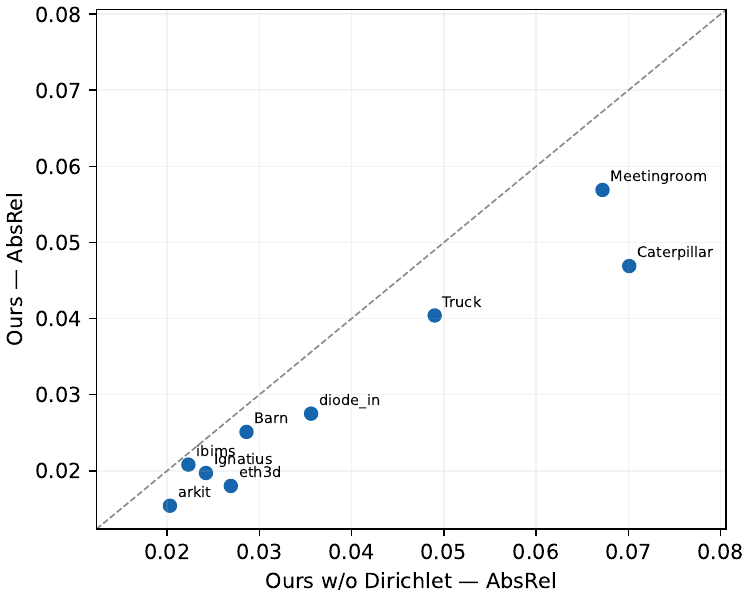}
\caption{Depth effect of residual completion; each point is a subset p50.}\label{fig:stages}
\end{figure}
\begin{table}[H]
\centering
\caption{Stage comparison at p50; MAE is in meters.}\label{tab:app-B-1}
\small
\setlength{\tabcolsep}{3pt}
\begin{tabular*}{\textwidth}{@{\extracolsep{\fill}}lrrrr@{}}
\toprule
Subset & Response AbsRel & Ours AbsRel & Response MAE & Ours MAE\\
\midrule
DIODE-in & 0.0356 & 0.0275 & 0.0792 & 0.0656\\
ETH3D & 0.0269 & 0.0180 & 0.1476 & 0.1082\\
iBims & 0.0223 & 0.0208 & 0.0604 & 0.0610\\
ARKit & 0.0203 & 0.0154 & 0.0256 & 0.0209\\
Barn & 0.0286 & 0.0251 & 0.2558 & 0.2258\\
Caterpillar & 0.0701 & 0.0469 & 0.3049 & 0.2369\\
Ignatius & 0.0242 & 0.0197 & 0.0652 & 0.0500\\
Meetingroom & 0.0672 & 0.0569 & 0.3409 & 0.2866\\
Truck & 0.0490 & 0.0404 & 0.1809 & 0.1471\\
\bottomrule
\end{tabular*}
\end{table}
All seven percentiles and failure counts are supplied in the accompanying CSVs.
\FloatBarrier

\clearpage
\section{Surface-Normal Quality}
\label{app:section-C}
\label{app:normal-quality}
\label{app:support}
\label{app:additional}

\subsection{Geometry of the complete method}

The complete Ours pipeline has lower macro median normal error than PriorDA, LDCM, Any2Full, and PromptDA. It has the lowest p50 NMed among these complete systems on seven of the nine subsets. Any2Full is lower on Caterpillar (21.1007\ensuremath{^\circ} versus 22.6456\ensuremath{^\circ}) and Truck (14.0926\ensuremath{^\circ} versus 14.5495\ensuremath{^\circ}). Tables~\ref{tab:app-C-1} and~\ref{tab:app-C-2} compare p50 normal errors; the distribution overview retains the lower- and higher-error cases.

Per-image mean angular error provides a complementary view of the surface, with greater influence from large angular deviations. Both normal statistics are reported at p50; neither is inferred from the other.

\subsection{Higher normal quality already present in the response stage}

The response-only output provides evidence of substantial further normal-quality potential within the same pipeline. Its p50 normal mean and median errors are lower than those of complete Ours on all nine subsets. For example, Meetingroom NMed changes from 19.3755\ensuremath{^\circ} in the final result to 13.1470\ensuremath{^\circ} before residual completion, and Truck from 14.5495\ensuremath{^\circ} to 9.9316\ensuremath{^\circ}. The response-only NMed is also lower than the four trained complete baselines on every subset.

This comparison identifies geometric accuracy already available before the final spatial solve. The complete pipeline's depth advantage remains important: its anchor agreement and lower AbsRel accompany spatially varying corrections to the calibrated field. Since those corrections alter local depth gradients, their propagation affects surface orientation as well as metric values. Preserving more of the response stage's normal quality is therefore a concrete direction for improving an already competitive complete reconstruction. The response-stage measurements provide a concrete normal-quality target for the spatial reconstruction stage.

Appendix~\ref{app:section-D} examines how the reconstruction weights affect this balance. Full percentile values are supplied in the accompanying CSVs.

\begin{table}[H]
\centering
\caption{P50 mean normal error (degrees).}\label{tab:app-C-1}
\small
\setlength{\tabcolsep}{3pt}
\begin{tabular*}{\textwidth}{@{\extracolsep{\fill}}lrrrrrr@{}}
\toprule
Subset & PriorDA & LDCM & Any2Full & PromptDA & Ours & \shortstack{Ours w/o\\Dirichlet}\\
\midrule
DIODE-in & 29.74 & 24.46 & 20.57 & 77.26 & 20.55 & 16.71\\
ETH3D & 37.69 & 27.07 & 24.93 & 67.47 & 23.84 & 22.28\\
iBims & 29.53 & 22.26 & 19.60 & 67.68 & 20.99 & 16.33\\
ARKit & 20.32 & 16.36 & 15.76 & 73.51 & 15.07 & 13.53\\
Barn & 25.86 & 19.04 & 20.45 & 65.62 & 19.48 & 15.84\\
Caterpillar & 41.48 & 41.17 & 32.02 & 74.94 & 34.37 & 27.77\\
Ignatius & 33.52 & 36.63 & 29.05 & 75.28 & 24.71 & 22.93\\
Meetingroom & 42.49 & 33.88 & 30.58 & 69.08 & 31.11 & 24.73\\
Truck & 33.13 & 26.59 & 24.26 & 66.56 & 27.07 & 19.98\\
\bottomrule
\end{tabular*}
\end{table}
\begin{table}[H]
\centering
\caption{P50 median normal error (degrees).}\label{tab:app-C-2}
\small
\setlength{\tabcolsep}{3pt}
\begin{tabular*}{\textwidth}{@{\extracolsep{\fill}}lrrrrrr@{}}
\toprule
Subset & PriorDA & LDCM & Any2Full & PromptDA & Ours & \shortstack{Ours w/o\\Dirichlet}\\
\midrule
DIODE-in & 19.63 & 13.04 & 12.30 & 77.59 & 10.05 & 9.10\\
ETH3D & 24.34 & 14.03 & 13.12 & 60.90 & 11.52 & 10.68\\
iBims & 18.87 & 11.67 & 11.26 & 65.96 & 10.83 & 8.63\\
ARKit & 12.74 & 9.40 & 10.05 & 71.84 & 8.90 & 8.45\\
Barn & 16.37 & 11.19 & 11.85 & 64.29 & 10.43 & 8.79\\
Caterpillar & 31.69 & 31.38 & 21.10 & 74.32 & 22.65 & 17.01\\
Ignatius & 27.80 & 29.13 & 24.60 & 73.05 & 18.05 & 17.43\\
Meetingroom & 34.14 & 23.33 & 19.98 & 66.68 & 19.38 & 13.15\\
Truck & 22.14 & 15.36 & 14.09 & 61.97 & 14.55 & 9.93\\
\bottomrule
\end{tabular*}
\end{table}
\begin{figure}[H]
\centering
\includegraphics[width=.75\textwidth]{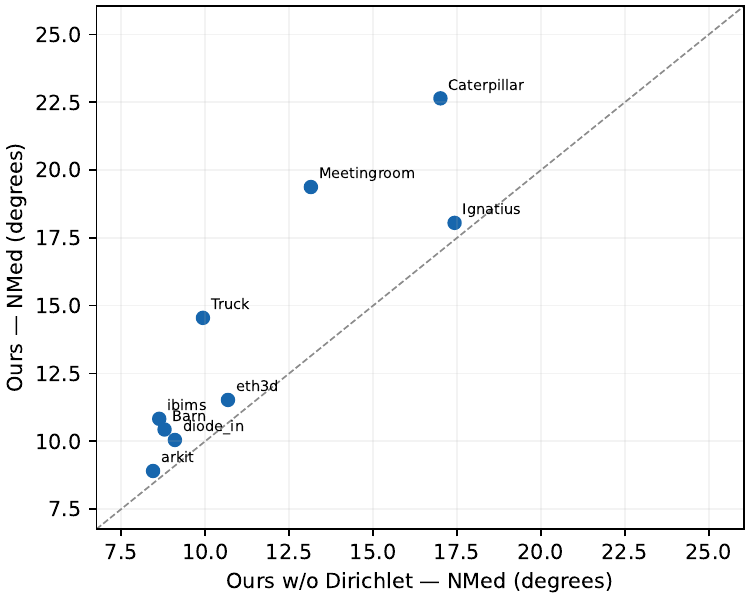}
\caption{Normal effect of residual completion; each point is a subset p50.}\label{fig:stage-normals}
\end{figure}
\FloatBarrier

\clearpage
\section{Residual-Reconstruction Weight Sensitivity}
\label{app:section-D}
\label{app:weights}

The residual objective combines gradient agreement, proximity to the calibrated prior, and Laplacian regularization. We retain $\lambda_{\rm grad}$=1 and examine the recorded data and Laplacian weights. The main experiments use ($\lambda_{\rm data}$,$\lambda_{\rm lap}$)=(10\ensuremath{^{-3}},10\ensuremath{^{-3}}), fixed independently of this sweep.

\Needspace{146pt}
\begingroup
\fontsize{8}{9.5}\selectfont
\setlength{\tabcolsep}{2pt}
\setlength{\LTcapwidth}{\textwidth}
\setlength{\LTpre}{5pt}
\setlength{\LTpost}{7pt}
\setlength{\appTableWidth}{\dimexpr\textwidth-12pt\relax}
\begin{longtable}{@{}>{\raggedright\arraybackslash}p{\dimexpr 0.250000\appTableWidth\relax}>{\raggedright\arraybackslash}p{\dimexpr 0.250000\appTableWidth\relax}>{\raggedright\arraybackslash}p{\dimexpr 0.250000\appTableWidth\relax}>{\raggedright\arraybackslash}p{\dimexpr 0.250000\appTableWidth\relax}@{}}
\caption{Recorded residual-reconstruction weight sweep. MAE is in meters and NMed in degrees.}\label{tab:app-D-1}\\
\toprule
$\lambda_{\rm data}$ & $\lambda_{\rm lap}$ & MAE (m) & NMed (deg)\\
\midrule
\endfirsthead
\multicolumn{4}{l}{Table \thetable\ (continued)}\\
\toprule
$\lambda_{\rm data}$ & $\lambda_{\rm lap}$ & MAE (m) & NMed (deg)\\
\midrule
\endhead
\midrule
\multicolumn{4}{r}{\scriptsize Continued on next page}\\
\endfoot
\bottomrule
\endlastfoot
1e-4 & 1e-4 & 0.1326 & 14.31\\
1e-4 & 1e-3 & 0.1326 & 14.31\\
1e-4 & 1e-2 & 0.1326 & 14.31\\
1e-3 & 1e-4 & 0.1336 & 14.04\\
1e-3 & 1e-3 & 0.1336 & 14.04\\
1e-3 & 1e-2 & 0.1335 & 14.04\\
1e-2 & 1e-4 & 0.1354 & 13.86\\
1e-2 & 1e-3 & 0.1354 & 13.86\\
1e-2 & 1e-2 & 0.1353 & 13.87\\
\end{longtable}
\endgroup

Increasing $\lambda_{\rm data}$ from 10\ensuremath{^{-4}} to 10\ensuremath{^{-2}} raises reported MAE from 0.1326 m to approximately 0.1353--0.1354 m while lowering NMed from 14.31\ensuremath{^\circ} to approximately 13.86--13.87\ensuremath{^\circ}. The stronger proximity penalty keeps the solution closer to the calibrated prior. Variation with $\lambda_{\rm lap}$ is much smaller over the tested range: the curves largely overlap at the recorded precision.

\begin{figure}[!ht]
\centering
\includegraphics[width=.78\textwidth]{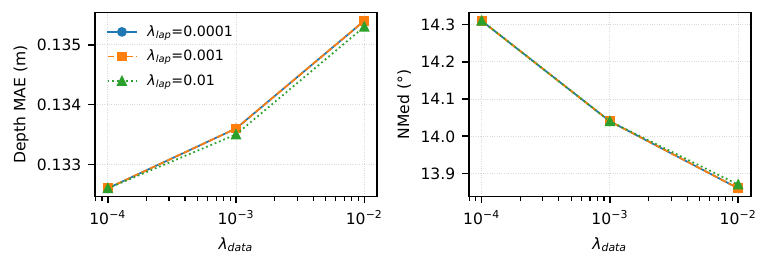}
\caption{Reconstruction weight sensitivity.}
\label{fig:sensitivity}
\end{figure}

This sweep shows that the backend can change depth and normal quality in different directions. The tested weights yield modest changes, while the response-stage comparison in Appendix~\ref{app:section-C} identifies a larger normal-quality difference.
\FloatBarrier

\clearpage
\section{Transfer across Monocular Priors}
\label{app:section-E}
\label{app:backbones}

We evaluate the same response and completion stages with MoGe-2, MoGe-2 Small~\citep{wang2025moge2}, and Depth Anything V2 Large and Base in addition to the MoGe-3 prior used in the main text. The comparison retains the nine subsets and the same incomplete-support evaluation. The response parameters adapt to the image and prior; the two-stage construction remains shared.

For every additional prior, residual completion lowers AbsRel on all nine subsets. The final depth error still depends on the prior: on ETH3D, for example, Ours obtains 0.0200 with MoGe-2, 0.0300 with MoGe-2 Small, 0.0256 with DAV2-L, and 0.0265 with DAV2-B. On DIODE-in, MoGe-2 yields 0.0272, close to the MoGe-3 result of 0.0275. These results show gains across priors alongside differences in their final accuracy.

The two grouped tables use backbones as columns and methods as rows; AbsRel and mean normal error (NMean, degrees) are paired under each backbone. Raw DAV2 entries that lack a defined metric scale remain marked as missing. Fixed-coordinate comparisons for every prior are consolidated in Appendix~\ref{app:section-G} so that the response's added freedom can be assessed against all three conventional choices.

\begin{table}[!ht]
\centering
\caption{Standard benchmarks: p50 comparison across monocular priors. NMean is mean normal error in degrees.}\label{tab:app-E-1}
\small
\setlength{\tabcolsep}{3pt}
\begin{tabular*}{\textwidth}{@{\extracolsep{\fill}}lrrrrrrrrrr@{}}
\toprule
 & \multicolumn{2}{c}{MoGe-3} & \multicolumn{2}{c}{MoGe-2} & \multicolumn{2}{c}{MoGe-2 S} & \multicolumn{2}{c}{DAV2-L} & \multicolumn{2}{c}{DAV2-B}\\
\cmidrule(lr){2-3}\cmidrule(lr){4-5}\cmidrule(lr){6-7}\cmidrule(lr){8-9}\cmidrule(lr){10-11}
Method & AbsRel & NMean & AbsRel & NMean & AbsRel & NMean & AbsRel & NMean & AbsRel & NMean\\
\midrule
\multicolumn{11}{l}{\textbf{DIODE-in}}\\
Raw & 0.2707 & 18.37 & 0.2789 & 19.04 & 0.2578 & 22.55 & -- & -- & -- & --\\
Ours w/o Dirichlet & 0.0356 & 16.71 & 0.0344 & 17.33 & 0.0493 & 21.18 & 0.0447 & 16.59 & 0.0454 & 17.22\\
Ours & 0.0275 & 20.55 & 0.0272 & 20.77 & 0.0345 & 24.27 & 0.0331 & 20.25 & 0.0339 & 20.57\\
\midrule
\multicolumn{11}{l}{\textbf{ETH3D}}\\
Raw & 0.0805 & 21.70 & 0.0839 & 22.83 & 0.1406 & 26.37 & -- & -- & -- & --\\
Ours w/o Dirichlet & 0.0269 & 22.28 & 0.0289 & 23.22 & 0.0447 & 27.36 & 0.0388 & 22.61 & 0.0436 & 23.12\\
Ours & 0.0180 & 23.84 & 0.0200 & 24.60 & 0.0300 & 28.20 & 0.0256 & 24.14 & 0.0265 & 24.68\\
\midrule
\multicolumn{11}{l}{\textbf{iBims}}\\
Raw & 0.0773 & 16.19 & 0.0717 & 17.06 & 0.0973 & 21.35 & -- & -- & -- & --\\
Ours w/o Dirichlet & 0.0223 & 16.33 & 0.0224 & 17.40 & 0.0358 & 21.83 & 0.0281 & 17.74 & 0.0309 & 18.34\\
Ours & 0.0208 & 20.99 & 0.0214 & 21.37 & 0.0276 & 24.76 & 0.0245 & 21.78 & 0.0248 & 22.35\\
\midrule
\multicolumn{11}{l}{\textbf{ARKit}}\\
Raw & 0.0678 & 12.97 & 0.0590 & 13.45 & 0.1263 & 17.37 & -- & -- & -- & --\\
Ours w/o Dirichlet & 0.0203 & 13.53 & 0.0205 & 13.88 & 0.0349 & 17.96 & 0.0309 & 14.65 & 0.0346 & 15.55\\
Ours & 0.0154 & 15.07 & 0.0154 & 15.28 & 0.0197 & 18.85 & 0.0192 & 15.91 & 0.0207 & 16.57\\
\bottomrule
\end{tabular*}
\end{table}

\begin{table}[!ht]
\centering
\caption{Tanks and Temples: p50 comparison across monocular priors. NMean is mean normal error in degrees.}\label{tab:app-E-2}
\small
\setlength{\tabcolsep}{3pt}
\begin{tabular*}{\textwidth}{@{\extracolsep{\fill}}lrrrrrrrrrr@{}}
\toprule
 & \multicolumn{2}{c}{MoGe-3} & \multicolumn{2}{c}{MoGe-2} & \multicolumn{2}{c}{MoGe-2 S} & \multicolumn{2}{c}{DAV2-L} & \multicolumn{2}{c}{DAV2-B}\\
\cmidrule(lr){2-3}\cmidrule(lr){4-5}\cmidrule(lr){6-7}\cmidrule(lr){8-9}\cmidrule(lr){10-11}
Method & AbsRel & NMean & AbsRel & NMean & AbsRel & NMean & AbsRel & NMean & AbsRel & NMean\\
\midrule
\multicolumn{11}{l}{\textbf{Barn}}\\
Raw & 0.0857 & 15.35 & 0.1025 & 16.85 & 0.1190 & 19.74 & -- & -- & -- & --\\
Ours w/o Dirichlet & 0.0286 & 15.84 & 0.0288 & 17.20 & 0.0402 & 20.23 & 0.0440 & 19.40 & 0.0430 & 19.57\\
Ours & 0.0251 & 19.48 & 0.0251 & 20.17 & 0.0313 & 22.45 & 0.0347 & 22.71 & 0.0335 & 22.55\\
\midrule
\multicolumn{11}{l}{\textbf{Caterpillar}}\\
Raw & 0.1393 & 27.52 & 0.1184 & 29.39 & 0.1054 & 36.54 & -- & -- & -- & --\\
Ours w/o Dirichlet & 0.0701 & 27.77 & 0.0673 & 29.42 & 0.0821 & 35.62 & 0.0604 & 30.29 & 0.0628 & 30.98\\
Ours & 0.0469 & 34.37 & 0.0473 & 35.55 & 0.0582 & 39.59 & 0.0473 & 35.80 & 0.0489 & 36.44\\
\midrule
\multicolumn{11}{l}{\textbf{Ignatius}}\\
Raw & 0.2643 & 21.93 & 0.2132 & 23.81 & 0.3487 & 30.41 & -- & -- & -- & --\\
Ours w/o Dirichlet & 0.0242 & 22.93 & 0.0246 & 24.45 & 0.0307 & 30.31 & 0.0275 & 25.20 & 0.0282 & 25.69\\
Ours & 0.0197 & 24.71 & 0.0208 & 25.79 & 0.0256 & 29.84 & 0.0229 & 25.78 & 0.0223 & 26.23\\
\midrule
\multicolumn{11}{l}{\textbf{Meetingroom}}\\
Raw & 0.1055 & 24.43 & 0.1151 & 27.13 & 0.1564 & 33.55 & -- & -- & -- & --\\
Ours w/o Dirichlet & 0.0672 & 24.73 & 0.0718 & 27.33 & 0.0971 & 33.92 & 0.0984 & 28.86 & 0.0980 & 29.88\\
Ours & 0.0569 & 31.11 & 0.0593 & 33.11 & 0.0714 & 38.30 & 0.0728 & 34.73 & 0.0748 & 35.49\\
\midrule
\multicolumn{11}{l}{\textbf{Truck}}\\
Raw & 0.0925 & 18.96 & 0.1121 & 21.39 & 0.1917 & 25.76 & -- & -- & -- & --\\
Ours w/o Dirichlet & 0.0490 & 19.98 & 0.0524 & 22.15 & 0.0683 & 26.54 & 0.0588 & 23.22 & 0.0601 & 23.76\\
Ours & 0.0404 & 27.07 & 0.0423 & 28.76 & 0.0527 & 31.66 & 0.0452 & 29.66 & 0.0473 & 29.96\\
\bottomrule
\end{tabular*}
\end{table}

Full percentile results and failure counts are provided in the accompanying CSVs.
\FloatBarrier

\clearpage
\section{Adaptive Response versus Fixed Coordinates}
\label{app:section-G}
\label{app:fixed-coordinates}

\subsection{Coordinate selection across priors and subsets}

GDisp, GMetric, and GLog restrict calibration to disparity, metric depth, and log-depth. The response-only variant adds the coordinate parameter while retaining one global affine relation in the selected response space. We compare these four outputs before interpreting the additional benefit of Dirichlet completion.

Across all five priors and nine subsets, response-only AbsRel is lower than the best of the three fixed-coordinate entries. This comparison covers 45 prior--subset combinations at p50. For each prior and subset, the reference is the best of GDisp, GMetric, and GLog.

For MoGe-3 on DIODE-in, the fixed AbsRel values are 0.0455 (GDisp), 0.0440 (GMetric), and 0.0416 (GLog), compared with 0.0356 for response-only and 0.0275 after completion. Appendix~\ref{app:section-E} shows that the same construction also improves the additional priors. The complete-method comparisons include the spatial backend, while response-only comparisons isolate the global calibration stage.

\subsection{Depth and normal behavior need separate comparisons}

Normal-error rankings vary across coordinates and subsets. For MoGe-3 on ETH3D, response-only p50 NMed is 10.6771\ensuremath{^\circ}, compared with 10.6612\ensuremath{^\circ} for GLog; on Meetingroom it is 13.1470\ensuremath{^\circ} versus 13.0269\ensuremath{^\circ}. With MoGe-2 Small, fixed coordinates also provide lower NMed on several subsets. 

The response is estimated for metric alignment, and its nonlinear gain changes the transfer of local geometry. Normal mean and median errors therefore provide separate evidence about that transfer. Appendix~\ref{app:section-C} compares the available normal quality with trained completion systems; this section compares it with the narrower fixed-coordinate model family.

\subsection{Subset comparisons and unsuccessful entries}

The new export contains all three fixed-coordinate configurations for all five priors. Tables~\ref{tab:app-G-1} and~\ref{tab:app-G-2} group the standard benchmarks and Tanks and Temples scenes, respectively. Within each subset block, backbones are columns and methods are rows, with paired AbsRel and mean normal error (NMean, degrees). Some newly completed configurations report failed entries; the accompanying CSVs retain their counts alongside the valid-score distributions. Reported percentiles summarize valid scores; failure counts are listed separately in the CSVs.

For example, DIODE-in GMetric with DAV2-B records 25 failures and GLog records 21. On ETH3D, the corresponding counts are 23 and 19.

\begin{table}[!ht]
\centering
\caption{Standard benchmarks: p50 comparison of adaptive and fixed responses. Each backbone has paired AbsRel and mean normal error (degrees).}\label{tab:app-G-1}
\small\setlength{\tabcolsep}{3pt}
\renewcommand{\arraystretch}{1.15}
\begin{tabular*}{\textwidth}{@{\extracolsep{\fill}}lrrrrrrrrrr@{}}
\toprule
 & \multicolumn{2}{c}{MoGe-3} & \multicolumn{2}{c}{MoGe-2} & \multicolumn{2}{c}{MoGe-2 S} & \multicolumn{2}{c}{DAV2-L} & \multicolumn{2}{c}{DAV2-B}\\
\cmidrule(lr){2-3}\cmidrule(lr){4-5}\cmidrule(lr){6-7}\cmidrule(lr){8-9}\cmidrule(lr){10-11}
Method & AbsRel & NMean & AbsRel & NMean & AbsRel & NMean & AbsRel & NMean & AbsRel & NMean\\
\midrule
\multicolumn{11}{l}{\textbf{DIODE}}\\
GDisp & 0.0455 & 16.86 & 0.0454 & 17.50 & 0.0596 & 21.26 & 0.0526 & 16.54 & 0.0529 & 17.28\\
GMetric & 0.0440 & 16.98 & 0.0443 & 17.48 & 0.0596 & 21.49 & 0.1519 & 30.82 & 0.1140 & 26.86\\
GLog & 0.0416 & 16.63 & 0.0415 & 17.30 & 0.0551 & 21.15 & 0.0821 & 20.88 & 0.0744 & 19.72\\
Ours w/o Dirichlet & 0.0356 & 16.71 & 0.0344 & 17.33 & 0.0493 & 21.18 & 0.0447 & 16.59 & 0.0454 & 17.22\\
Ours & 0.0275 & 20.55 & 0.0272 & 20.77 & 0.0345 & 24.27 & 0.0331 & 20.25 & 0.0339 & 20.57\\
\midrule
\multicolumn{11}{l}{\textbf{ETH3D}}\\
GDisp & 0.0348 & 22.38 & 0.0346 & 23.41 & 0.0536 & 27.44 & 0.0429 & 22.45 & 0.0457 & 23.14\\
GMetric & 0.0328 & 22.24 & 0.0351 & 23.23 & 0.0534 & 27.12 & 0.2008 & 37.36 & 0.1740 & 34.97\\
GLog & 0.0318 & 22.05 & 0.0332 & 23.12 & 0.0496 & 27.02 & 0.0780 & 25.60 & 0.0729 & 25.07\\
Ours w/o Dirichlet & 0.0269 & 22.28 & 0.0289 & 23.22 & 0.0447 & 27.36 & 0.0388 & 22.61 & 0.0436 & 23.12\\
Ours & 0.0180 & 23.84 & 0.0200 & 24.60 & 0.0300 & 28.20 & 0.0256 & 24.14 & 0.0265 & 24.68\\
\midrule
\multicolumn{11}{l}{\textbf{iBims}}\\
GDisp & 0.0292 & 16.53 & 0.0274 & 17.26 & 0.0398 & 21.71 & 0.0318 & 17.37 & 0.0353 & 18.05\\
GMetric & 0.0244 & 16.49 & 0.0246 & 17.34 & 0.0386 & 21.76 & 0.1707 & 33.50 & 0.1284 & 30.16\\
GLog & 0.0256 & 16.27 & 0.0255 & 17.12 & 0.0387 & 21.72 & 0.0796 & 22.80 & 0.0664 & 22.21\\
Ours w/o Dirichlet & 0.0223 & 16.33 & 0.0224 & 17.40 & 0.0358 & 21.83 & 0.0281 & 17.74 & 0.0309 & 18.34\\
Ours & 0.0208 & 20.99 & 0.0214 & 21.37 & 0.0276 & 24.76 & 0.0245 & 21.78 & 0.0248 & 22.35\\
\midrule
\multicolumn{11}{l}{\textbf{ARKit}}\\
GDisp & 0.0227 & 13.41 & 0.0228 & 13.83 & 0.0380 & 17.98 & 0.0336 & 14.49 & 0.0378 & 15.39\\
GMetric & 0.0227 & 13.38 & 0.0228 & 13.80 & 0.0384 & 17.86 & 0.1193 & 29.75 & 0.1013 & 27.15\\
GLog & 0.0221 & 13.27 & 0.0220 & 13.72 & 0.0378 & 17.83 & 0.0635 & 19.46 & 0.0599 & 18.92\\
Ours w/o Dirichlet & 0.0203 & 13.53 & 0.0205 & 13.88 & 0.0349 & 17.96 & 0.0309 & 14.65 & 0.0346 & 15.55\\
Ours & 0.0154 & 15.07 & 0.0154 & 15.28 & 0.0197 & 18.85 & 0.0192 & 15.91 & 0.0207 & 16.57\\
\bottomrule
\end{tabular*}
\end{table}

\begin{table}[!ht]
\centering
\caption{Tanks and Temples: p50 comparison of adaptive and fixed responses. Each backbone has paired AbsRel and mean normal error (degrees).}\label{tab:app-G-2}
\small\setlength{\tabcolsep}{3pt}
\renewcommand{\arraystretch}{1.15}
\begin{tabular*}{\textwidth}{@{\extracolsep{\fill}}lrrrrrrrrrr@{}}
\toprule
 & \multicolumn{2}{c}{MoGe-3} & \multicolumn{2}{c}{MoGe-2} & \multicolumn{2}{c}{MoGe-2 S} & \multicolumn{2}{c}{DAV2-L} & \multicolumn{2}{c}{DAV2-B}\\
\cmidrule(lr){2-3}\cmidrule(lr){4-5}\cmidrule(lr){6-7}\cmidrule(lr){8-9}\cmidrule(lr){10-11}
Method & AbsRel & NMean & AbsRel & NMean & AbsRel & NMean & AbsRel & NMean & AbsRel & NMean\\
\midrule
\multicolumn{11}{l}{\textbf{Barn}}\\
GDisp & 0.0443 & 15.99 & 0.0441 & 17.28 & 0.0514 & 19.96 & 0.0642 & 19.23 & 0.0633 & 19.21\\
GMetric & 0.1233 & 24.54 & 0.1256 & 25.85 & 0.1169 & 24.99 & 0.2057 & 39.09 & 0.2015 & 39.06\\
GLog & 0.0670 & 17.68 & 0.0703 & 18.83 & 0.0693 & 20.74 & 0.1056 & 21.99 & 0.1066 & 22.07\\
Ours w/o Dirichlet & 0.0286 & 15.84 & 0.0288 & 17.20 & 0.0402 & 20.23 & 0.0440 & 19.40 & 0.0430 & 19.57\\
Ours & 0.0251 & 19.48 & 0.0251 & 20.17 & 0.0313 & 22.45 & 0.0347 & 22.71 & 0.0335 & 22.55\\
\midrule
\multicolumn{11}{l}{\textbf{Caterpillar}}\\
GDisp & 0.0936 & 28.82 & 0.0900 & 30.11 & 0.0963 & 35.17 & 0.0937 & 30.62 & 0.0957 & 31.51\\
GMetric & 0.1819 & 43.17 & 0.1784 & 42.32 & 0.1771 & 42.38 & 0.1957 & 45.15 & 0.1969 & 45.39\\
GLog & 0.1379 & 34.51 & 0.1309 & 34.25 & 0.1283 & 36.59 & 0.1468 & 36.75 & 0.1507 & 37.82\\
Ours w/o Dirichlet & 0.0701 & 27.77 & 0.0673 & 29.42 & 0.0821 & 35.62 & 0.0604 & 30.29 & 0.0628 & 30.98\\
Ours & 0.0469 & 34.37 & 0.0473 & 35.55 & 0.0582 & 39.59 & 0.0473 & 35.80 & 0.0489 & 36.44\\
\midrule
\multicolumn{11}{l}{\textbf{Ignatius}}\\
GDisp & 0.0461 & 34.34 & 0.0453 & 34.48 & 0.0468 & 36.55 & 0.0457 & 34.24 & 0.0466 & 34.67\\
GMetric & 0.0550 & 42.09 & 0.0552 & 42.03 & 0.0555 & 42.35 & 0.0568 & 42.46 & 0.0575 & 42.51\\
GLog & 0.0517 & 39.26 & 0.0515 & 39.14 & 0.0520 & 39.98 & 0.0527 & 39.52 & 0.0532 & 39.84\\
Ours w/o Dirichlet & 0.0242 & 22.93 & 0.0246 & 24.45 & 0.0307 & 30.31 & 0.0275 & 25.20 & 0.0282 & 25.69\\
Ours & 0.0197 & 24.71 & 0.0208 & 25.79 & 0.0256 & 29.84 & 0.0229 & 25.78 & 0.0223 & 26.23\\
\midrule
\multicolumn{11}{l}{\textbf{Meetingroom}}\\
GDisp & 0.0812 & 24.66 & 0.0850 & 27.28 & 0.1102 & 34.37 & 0.1058 & 29.02 & 0.1047 & 30.01\\
GMetric & 0.0791 & 24.78 & 0.0833 & 27.34 & 0.1079 & 33.69 & 0.4391 & 48.61 & 0.3224 & 43.37\\
GLog & 0.0752 & 24.49 & 0.0790 & 27.14 & 0.1018 & 33.75 & 0.1990 & 33.38 & 0.1720 & 33.18\\
Ours w/o Dirichlet & 0.0672 & 24.73 & 0.0718 & 27.33 & 0.0971 & 33.92 & 0.0984 & 28.86 & 0.0980 & 29.88\\
Ours & 0.0569 & 31.11 & 0.0593 & 33.11 & 0.0714 & 38.30 & 0.0728 & 34.73 & 0.0748 & 35.49\\
\midrule
\multicolumn{11}{l}{\textbf{Truck}}\\
GDisp & 0.0740 & 21.15 & 0.0760 & 23.21 & 0.0850 & 26.91 & 0.0883 & 24.14 & 0.0897 & 24.57\\
GMetric & 0.1920 & 42.96 & 0.1927 & 43.32 & 0.1876 & 42.04 & 0.2318 & 51.51 & 0.2340 & 51.82\\
GLog & 0.1197 & 27.42 & 0.1194 & 28.62 & 0.1207 & 30.63 & 0.1536 & 35.23 & 0.1550 & 35.54\\
Ours w/o Dirichlet & 0.0490 & 19.98 & 0.0524 & 22.15 & 0.0683 & 26.54 & 0.0588 & 23.22 & 0.0601 & 23.76\\
Ours & 0.0404 & 27.07 & 0.0423 & 28.76 & 0.0527 & 31.66 & 0.0452 & 29.66 & 0.0473 & 29.96\\
\bottomrule
\end{tabular*}
\end{table}

\FloatBarrier

\clearpage
\section{Reference Geometry and PromptDA Diagnostics}
\label{app:section-H}
\label{app:reference-diagnostics}

\subsection{Benchmark inclusion, support structure, and depth range}
\label{app:tt-selection}
\label{app:reference-support}

The benchmark combines accurate reference geometry with heterogeneous scene content and support. DIODE indoor and ETH3D supply laser-based reference geometry; iBims-1 supplies reference depth from precise scans; the selected ARKitScenes stream uses FARO reference; and Tanks and Temples depth is projected from laser geometry into calibrated views. The sample includes broadly covered indoor scenes and object-centered scans with localized valid support. The additional observation removal operates on these existing differences.

\textbf{Why five Tanks and Temples scenes.} Barn, Caterpillar, Ignatius, Meetingroom, and Truck satisfy the common calibration and reference-support requirements. The projection pipeline uses one public COLMAP/OPENCV calibration bundle, TNT\_GOF. Church has no calibration in that bundle and is omitted to maintain a common projection convention.

Courthouse was processed but excluded after auditing its projected reference across 1,106 frames. Reprojection of the source cloud agreed with the cached rendered depth to a median discrepancy of approximately 4--9 mm in the reported checks, and point-cloud color reprojection supported the image--pose correspondence. The remaining problem was concentrated around vegetation: scan rays passing through gaps in foliage can assign distant background depths to foreground image regions. Using depths greater than twice each frame's median as the far-pixel criterion, 20.8\% of frames had more than 10\% far pixels. Full-resolution minimum-depth rendering followed by min-pooling at radii 4, 8, and 16 recovered approximately 23--57\% of these pixels. Correcting the remaining mismatch would require substantial scene-specific filtering, so Courthouse is excluded from the common benchmark.

The five retained scenes thus share the same calibration source and projection procedure. The coverage of the retained support is allowed to vary; low coverage alone is not an exclusion criterion. Dataset sampling follows original image indices without coverage-based prefiltering. The retained counts are shown below, totaling 3,698 unique images and 11,094 image--observation-case entries before method-specific validity filtering. ARKit uses at most four uniformly selected frames per video at 384\ensuremath{\times}512; ARKit and iBims reference depth is capped at 10 m.

\textbf{Areal support, holes, and connectivity.} Valid coverage is the fraction of image pixels with valid reference depth. The interior-hole fraction is the area of filled interior holes divided by the support area after hole filling. The largest-component fraction is the largest connected valid component divided by the total number of valid pixels. Tables~\ref{tab:app-H-1}--\ref{tab:app-H-3} each report one percentile (p16, p50, or p84) of these statistics together. Coverage, holes, and connectivity use per-image percentiles; metric depth uses pooled valid pixels. The image and pixel counts are labeled separately.

\Needspace{146pt}
\begingroup
\fontsize{8}{9.5}\selectfont
\setlength{\tabcolsep}{2pt}
\setlength{\LTcapwidth}{\textwidth}
\setlength{\LTpre}{5pt}
\setlength{\LTpost}{7pt}
\setlength{\appTableWidth}{\dimexpr\textwidth-24pt\relax}
\begin{longtable}{@{}>{\raggedright\arraybackslash}p{\dimexpr 0.166667\appTableWidth\relax}>{\raggedright\arraybackslash}p{\dimexpr 0.083333\appTableWidth\relax}>{\raggedright\arraybackslash}p{\dimexpr 0.156250\appTableWidth\relax}>{\raggedright\arraybackslash}p{\dimexpr 0.135417\appTableWidth\relax}>{\raggedright\arraybackslash}p{\dimexpr 0.135417\appTableWidth\relax}>{\raggedright\arraybackslash}p{\dimexpr 0.177083\appTableWidth\relax}>{\raggedright\arraybackslash}p{\dimexpr 0.145833\appTableWidth\relax}@{}}
\caption{P16 reference-geometry statistics.}\label{tab:app-H-1}\\
\toprule
Subset & Images & Valid pixels & Valid fraction & Hole fraction & Largest comp. & Depth (m)\\
\midrule
\endfirsthead
\multicolumn{7}{l}{Table \thetable\ (continued)}\\
\toprule
Subset & Images & Valid pixels & Valid fraction & Hole fraction & Largest comp. & Depth (m)\\
\midrule
\endhead
\midrule
\multicolumn{7}{r}{\scriptsize Continued on next page}\\
\endfoot
\bottomrule
\endlastfoot
DIODE-in & 325 & 227970748 & .8270 & .0002 & .9940 & 1.523\\
\addlinespace[3pt]
ETH3D & 454 & 459188168 & .4926 & .0313 & .3470 & 2.075\\
\addlinespace[3pt]
iBims & 100 & 28530051 & .8479 & .0006 & .9997 & 1.633\\
\addlinespace[3pt]
ARKit & 1141 & 197039280 & .7589 & .0013 & .9141 & .675\\
\addlinespace[3pt]
Barn & 410 & 104872017 & .2874 & .0000 & .9999 & 3.777\\
\addlinespace[3pt]
Caterpillar & 383 & 71222396 & .2145 & .0191 & .9994 & 2.378\\
\addlinespace[3pt]
Ignatius & 263 & 23208853 & .0989 & .0000 & .9995 & 1.641\\
\addlinespace[3pt]
Meetingroom & 371 & 185650467 & .9430 & .0070 & .9982 & 2.721\\
\addlinespace[3pt]
Truck & 251 & 67868863 & .3663 & .0051 & .9997 & 2.011\\
\end{longtable}
\endgroup

\Needspace{146pt}
\begingroup
\fontsize{8}{9.5}\selectfont
\setlength{\tabcolsep}{2pt}
\setlength{\LTcapwidth}{\textwidth}
\setlength{\LTpre}{5pt}
\setlength{\LTpost}{7pt}
\setlength{\appTableWidth}{\dimexpr\textwidth-24pt\relax}
\begin{longtable}{@{}>{\raggedright\arraybackslash}p{\dimexpr 0.166667\appTableWidth\relax}>{\raggedright\arraybackslash}p{\dimexpr 0.083333\appTableWidth\relax}>{\raggedright\arraybackslash}p{\dimexpr 0.156250\appTableWidth\relax}>{\raggedright\arraybackslash}p{\dimexpr 0.135417\appTableWidth\relax}>{\raggedright\arraybackslash}p{\dimexpr 0.135417\appTableWidth\relax}>{\raggedright\arraybackslash}p{\dimexpr 0.177083\appTableWidth\relax}>{\raggedright\arraybackslash}p{\dimexpr 0.145833\appTableWidth\relax}@{}}
\caption{P50 reference-geometry statistics.}\label{tab:app-H-2}\\
\toprule
Subset & Images & Valid pixels & Valid fraction & Hole fraction & Largest comp. & Depth (m)\\
\midrule
\endfirsthead
\multicolumn{7}{l}{Table \thetable\ (continued)}\\
\toprule
Subset & Images & Valid pixels & Valid fraction & Hole fraction & Largest comp. & Depth (m)\\
\midrule
\endhead
\midrule
\multicolumn{7}{r}{\scriptsize Continued on next page}\\
\endfoot
\bottomrule
\endlastfoot
DIODE-in & 325 & 227970748 & .9938 & .0025 & 1.0000 & 2.386\\
\addlinespace[3pt]
ETH3D & 454 & 459188168 & .6801 & .0900 & .5816 & 3.943\\
\addlinespace[3pt]
iBims & 100 & 28530051 & .9656 & .0100 & 1.0000 & 2.650\\
\addlinespace[3pt]
ARKit & 1141 & 197039280 & .9056 & .0164 & .9953 & 1.198\\
\addlinespace[3pt]
Barn & 410 & 104872017 & .4565 & .0001 & 1.0000 & 5.964\\
\addlinespace[3pt]
Caterpillar & 383 & 71222396 & .3469 & .0362 & .9998 & 3.426\\
\addlinespace[3pt]
Ignatius & 263 & 23208853 & .1554 & .0000 & .9999 & 2.404\\
\addlinespace[3pt]
Meetingroom & 371 & 185650467 & .9686 & .0244 & .9996 & 4.943\\
\addlinespace[3pt]
Truck & 251 & 67868863 & .5300 & .0146 & 1.0000 & 2.989\\
\end{longtable}
\endgroup

\Needspace{146pt}
\begingroup
\fontsize{8}{9.5}\selectfont
\setlength{\tabcolsep}{2pt}
\setlength{\LTcapwidth}{\textwidth}
\setlength{\LTpre}{5pt}
\setlength{\LTpost}{7pt}
\setlength{\appTableWidth}{\dimexpr\textwidth-24pt\relax}
\begin{longtable}{@{}>{\raggedright\arraybackslash}p{\dimexpr 0.166667\appTableWidth\relax}>{\raggedright\arraybackslash}p{\dimexpr 0.083333\appTableWidth\relax}>{\raggedright\arraybackslash}p{\dimexpr 0.156250\appTableWidth\relax}>{\raggedright\arraybackslash}p{\dimexpr 0.135417\appTableWidth\relax}>{\raggedright\arraybackslash}p{\dimexpr 0.135417\appTableWidth\relax}>{\raggedright\arraybackslash}p{\dimexpr 0.177083\appTableWidth\relax}>{\raggedright\arraybackslash}p{\dimexpr 0.145833\appTableWidth\relax}@{}}
\caption{P84 reference-geometry statistics.}\label{tab:app-H-3}\\
\toprule
Subset & Images & Valid pixels & Valid fraction & Hole fraction & Largest comp. & Depth (m)\\
\midrule
\endfirsthead
\multicolumn{7}{l}{Table \thetable\ (continued)}\\
\toprule
Subset & Images & Valid pixels & Valid fraction & Hole fraction & Largest comp. & Depth (m)\\
\midrule
\endhead
\midrule
\multicolumn{7}{r}{\scriptsize Continued on next page}\\
\endfoot
\bottomrule
\endlastfoot
DIODE-in & 325 & 227970748 & .9998 & .0197 & 1.0000 & 4.485\\
\addlinespace[3pt]
ETH3D & 454 & 459188168 & .8379 & .1701 & .8826 & 11.359\\
\addlinespace[3pt]
iBims & 100 & 28530051 & .9981 & .0805 & 1.0000 & 4.468\\
\addlinespace[3pt]
ARKit & 1141 & 197039280 & .9799 & .0536 & .9999 & 1.932\\
\addlinespace[3pt]
Barn & 410 & 104872017 & .7183 & .0018 & 1.0000 & 10.208\\
\addlinespace[3pt]
Caterpillar & 383 & 71222396 & .5187 & .0504 & 1.0000 & 5.363\\
\addlinespace[3pt]
Ignatius & 263 & 23208853 & .2306 & .0001 & 1.0000 & 3.311\\
\addlinespace[3pt]
Meetingroom & 371 & 185650467 & .9883 & .0488 & .9999 & 8.253\\
\addlinespace[3pt]
Truck & 251 & 67868863 & .6589 & .0219 & 1.0000 & 4.363\\
\end{longtable}
\endgroup

Ignatius illustrates why connectivity and coverage are different quantities. At the median, valid depth covers only 15.54\% of the image, while 99.99\% of the valid pixels belong to one component and the interior-hole fraction is numerically zero. Barn and Caterpillar also have substantially incomplete coverage with highly connected valid support. Such support can preserve an object's shape while occupying only a limited part of the image.

ETH3D exhibits a different structure: median coverage is 68.01\%, its largest component contains 58.16\% of valid pixels, and the interior-hole fraction is 9.00\%. Its support is therefore more fragmented despite covering a larger fraction of the image than Ignatius. Connectivity and hole statistics characterize the reference support; normal errors are evaluated using the geometric validity masks.

The depth distributions also span different operating scales. ARKit's pooled p50 depth is 1.198 m, compared with 5.964 m in Barn. ETH3D extends to 15.967 m at p93, while Ignatius reaches 3.822 m. These measurements describe the reference geometry before further observation removal; the removed image-area fraction is not the remaining valid-anchor fraction.

\subsection{PromptDA inference protocol and support diagnostic}
\label{app:promptda}

We use the released Prompt-Depth-Anything-Large (ViT-L) checkpoint and inference interface, following its RGB resizing rule with a maximum side of 1008 pixels and dimensions divisible by 14. The nominal prompt stride of 7.5 matches the spatial ratio of the released ARKit setting. Prompt normalization and output denormalization remain inside the model. Reference-derived prompts contain no added noise and use the same deterministic observation masks as the other methods.

PromptDA's error distribution contains low-error cases on several broadly covered subsets despite a near-unit median AbsRel. On ARKit, p16 AbsRel is 0.0245 before p50 reaches 0.9876. DIODE and Meetingroom show the same qualitative transition.

\Needspace{98pt}
\begingroup
\fontsize{8}{9.5}\selectfont
\setlength{\tabcolsep}{2pt}
\setlength{\LTcapwidth}{\textwidth}
\setlength{\LTpre}{5pt}
\setlength{\LTpost}{7pt}
\setlength{\appTableWidth}{\dimexpr\textwidth-32pt\relax}
\begin{longtable}{@{}>{\raggedright\arraybackslash}p{\dimexpr 0.150000\appTableWidth\relax}>{\raggedright\arraybackslash}p{\dimexpr 0.130000\appTableWidth\relax}>{\raggedright\arraybackslash}p{\dimexpr 0.080000\appTableWidth\relax}>{\raggedright\arraybackslash}p{\dimexpr 0.080000\appTableWidth\relax}>{\raggedright\arraybackslash}p{\dimexpr 0.100000\appTableWidth\relax}>{\raggedright\arraybackslash}p{\dimexpr 0.100000\appTableWidth\relax}>{\raggedright\arraybackslash}p{\dimexpr 0.100000\appTableWidth\relax}>{\raggedright\arraybackslash}p{\dimexpr 0.100000\appTableWidth\relax}>{\raggedright\arraybackslash}p{\dimexpr 0.160000\appTableWidth\relax}@{}}
\caption{P16 comparison.  All four metrics use the same methods and subsets. Failure counts are listed in the order AbsRel / MAE / normal mean / normal median.}\label{tab:app-H-4}\\
\toprule
Subset & Method & Valid p50 & LCC p50 & AbsRel & MAE (m) & Normal mean ($^\circ$) & Normal median ($^\circ$) & Failures\\
\midrule
\endfirsthead
\multicolumn{9}{l}{Table \thetable\ (continued)}\\
\toprule
Subset & Method & Valid p50 & LCC p50 & AbsRel & MAE (m) & Normal mean ($^\circ$) & Normal median ($^\circ$) & Failures\\
\midrule
\endhead
\midrule
\multicolumn{9}{r}{\scriptsize Continued on next page}\\
\endfoot
\bottomrule
\endlastfoot
diode\_in & PromptDA & 0.9938 & 1.0000 & 0.0226 & 0.0567 & 20.5383 & 11.9354 & 0/0/0/0\\
\addlinespace[3pt]
arkit & PromptDA & 0.9056 & 0.9953 & 0.0245 & 0.0314 & 17.2584 & 10.1660 & 0/0/3/3\\
\addlinespace[3pt]
Meetingroom & PromptDA & 0.9686 & 0.9996 & 0.0565 & 0.2533 & 30.7781 & 19.3951 & 0/0/0/0\\
\addlinespace[3pt]
Ignatius & PromptDA & 0.1554 & 0.9999 & 0.6307 & 1.5352 & 64.1742 & 48.4493 & 0/0/0/0\\
\end{longtable}
\endgroup

\Needspace{98pt}
\begingroup
\fontsize{8}{9.5}\selectfont
\setlength{\tabcolsep}{2pt}
\setlength{\LTcapwidth}{\textwidth}
\setlength{\LTpre}{5pt}
\setlength{\LTpost}{7pt}
\setlength{\appTableWidth}{\dimexpr\textwidth-32pt\relax}
\begin{longtable}{@{}>{\raggedright\arraybackslash}p{\dimexpr 0.150000\appTableWidth\relax}>{\raggedright\arraybackslash}p{\dimexpr 0.130000\appTableWidth\relax}>{\raggedright\arraybackslash}p{\dimexpr 0.080000\appTableWidth\relax}>{\raggedright\arraybackslash}p{\dimexpr 0.080000\appTableWidth\relax}>{\raggedright\arraybackslash}p{\dimexpr 0.100000\appTableWidth\relax}>{\raggedright\arraybackslash}p{\dimexpr 0.100000\appTableWidth\relax}>{\raggedright\arraybackslash}p{\dimexpr 0.100000\appTableWidth\relax}>{\raggedright\arraybackslash}p{\dimexpr 0.100000\appTableWidth\relax}>{\raggedright\arraybackslash}p{\dimexpr 0.160000\appTableWidth\relax}@{}}
\caption{P50 comparison.  All four metrics use the same methods and subsets. Failure counts are listed in the order AbsRel / MAE / normal mean / normal median.}\label{tab:app-H-5}\\
\toprule
Subset & Method & Valid p50 & LCC p50 & AbsRel & MAE (m) & Normal mean ($^\circ$) & Normal median ($^\circ$) & Failures\\
\midrule
\endfirsthead
\multicolumn{9}{l}{Table \thetable\ (continued)}\\
\toprule
Subset & Method & Valid p50 & LCC p50 & AbsRel & MAE (m) & Normal mean ($^\circ$) & Normal median ($^\circ$) & Failures\\
\midrule
\endhead
\midrule
\multicolumn{9}{r}{\scriptsize Continued on next page}\\
\endfoot
\bottomrule
\endlastfoot
diode\_in & PromptDA & 0.9938 & 1.0000 & 0.9943 & 2.0260 & 77.2586 & 77.5938 & 0/0/0/0\\
\addlinespace[3pt]
arkit & PromptDA & 0.9056 & 0.9953 & 0.9876 & 0.9078 & 73.5105 & 71.8400 & 0/0/3/3\\
\addlinespace[3pt]
Meetingroom & PromptDA & 0.9686 & 0.9996 & 0.9917 & 4.8201 & 69.0824 & 66.6804 & 0/0/0/0\\
\addlinespace[3pt]
Ignatius & PromptDA & 0.1554 & 0.9999 & 0.9983 & 2.4738 & 75.2779 & 73.0478 & 0/0/0/0\\
\end{longtable}
\endgroup

\Needspace{98pt}
\begingroup
\fontsize{8}{9.5}\selectfont
\setlength{\tabcolsep}{2pt}
\setlength{\LTcapwidth}{\textwidth}
\setlength{\LTpre}{5pt}
\setlength{\LTpost}{7pt}
\setlength{\appTableWidth}{\dimexpr\textwidth-32pt\relax}
\begin{longtable}{@{}>{\raggedright\arraybackslash}p{\dimexpr 0.150000\appTableWidth\relax}>{\raggedright\arraybackslash}p{\dimexpr 0.130000\appTableWidth\relax}>{\raggedright\arraybackslash}p{\dimexpr 0.080000\appTableWidth\relax}>{\raggedright\arraybackslash}p{\dimexpr 0.080000\appTableWidth\relax}>{\raggedright\arraybackslash}p{\dimexpr 0.100000\appTableWidth\relax}>{\raggedright\arraybackslash}p{\dimexpr 0.100000\appTableWidth\relax}>{\raggedright\arraybackslash}p{\dimexpr 0.100000\appTableWidth\relax}>{\raggedright\arraybackslash}p{\dimexpr 0.100000\appTableWidth\relax}>{\raggedright\arraybackslash}p{\dimexpr 0.160000\appTableWidth\relax}@{}}
\caption{P84 comparison.  All four metrics use the same methods and subsets. Failure counts are listed in the order AbsRel / MAE / normal mean / normal median.}\label{tab:app-H-6}\\
\toprule
Subset & Method & Valid p50 & LCC p50 & AbsRel & MAE (m) & Normal mean ($^\circ$) & Normal median ($^\circ$) & Failures\\
\midrule
\endfirsthead
\multicolumn{9}{l}{Table \thetable\ (continued)}\\
\toprule
Subset & Method & Valid p50 & LCC p50 & AbsRel & MAE (m) & Normal mean ($^\circ$) & Normal median ($^\circ$) & Failures\\
\midrule
\endhead
\midrule
\multicolumn{9}{r}{\scriptsize Continued on next page}\\
\endfoot
\bottomrule
\endlastfoot
diode\_in & PromptDA & 0.9938 & 1.0000 & 0.9975 & 4.0551 & 89.2484 & 90.3284 & 0/0/0/0\\
\addlinespace[3pt]
arkit & PromptDA & 0.9056 & 0.9953 & 0.9963 & 1.6984 & 87.6166 & 88.3735 & 0/0/3/3\\
\addlinespace[3pt]
Meetingroom & PromptDA & 0.9686 & 0.9996 & 0.9956 & 6.7648 & 73.4394 & 72.4674 & 0/0/0/0\\
\addlinespace[3pt]
Ignatius & PromptDA & 0.1554 & 0.9999 & 0.9988 & 3.6442 & 86.6829 & 88.2056 & 0/0/0/0\\
\end{longtable}
\endgroup

\textbf{Tables~\ref{tab:app-H-4}--\ref{tab:app-H-6}.} PromptDA at p16, p50, and p84 on the diagnostic subsets, with the same four error metrics at each percentile. Coverage and connectivity columns remain the per-image p50 support statistics, explicitly labeled as such.

Ignatius remains difficult even at the low-error end: p16 AbsRel is 0.6307, compared with 0.9983 at p50. Its nearly connected support occupies only a small part of the image. In contrast, ARKit and Meetingroom have similarly high connectivity but much broader coverage. This pattern is consistent with sensitivity to incomplete spatial support. Appendix~\ref{app:section-F} includes PromptDA in the complete p50 comparison; the supplied CSVs retain all seven percentiles and failure counts.

The remaining percentiles (p7, p31, p69, and p93), together with all seven exported percentiles and failure counts, are provided in the accompanying CSVs: \nolinkurl{table3_depth_absrel.csv}, \nolinkurl{table4_depth_mae.csv}, \nolinkurl{table1_normal_mean.csv}, \nolinkurl{table2_normal_median.csv}.
The full seven-percentile reference statistics are provided in \nolinkurl{reference_geometry.csv}.
\FloatBarrier

\end{document}